\pdfoutput=1
\PassOptionsToPackage{table}{xcolor}
\documentclass[11pt]{article}

\usepackage[T1]{fontenc}
\usepackage[utf8]{inputenc}
\usepackage{lmodern}
\usepackage[a4paper,margin=1in]{geometry}
\usepackage{amsmath,amssymb}
\usepackage{booktabs}
\usepackage{graphicx}
\usepackage{tabularx}
\usepackage{multirow}
\usepackage{float}
\usepackage{xcolor}
\usepackage{colortbl}
\usepackage{enumitem}
\usepackage{csquotes}
\usepackage{microtype}
\usepackage{algorithm}
\usepackage{algpseudocode}
\usepackage[section]{placeins}
\usepackage{natbib}
\setcitestyle{aysep={}}
\usepackage{hyperref}
\hypersetup{colorlinks=true,linkcolor=blue!50!black,citecolor=blue!50!black,urlcolor=blue!50!black}
\renewcommand{\cite}{\citep}

\graphicspath{{figures/}}

\definecolor{gain}{RGB}{200,255,200}
\definecolor{loss}{RGB}{255,200,200}
\definecolor{lightgray}{gray}{0.9}

\title{Interpretive Variation and Discourse Structure in Detecting Enthymemes in Controversial Political Discourse}

\author{Martial Pastor\thanks{Corresponding author.} \qquad Nelleke Oostdijk\\[6pt]
Centre for Language Studies, Radboud University, Nijmegen, The Netherlands\\
\texttt{martial.pastor@ru.nl} \quad \texttt{nelleke.oostdijk@ru.nl}}

\date{}

\begin{document}

\maketitle

\begin{abstract}
Enthymemes are arguments that leave a premise or conclusion unstated, and they are a common persuasion device on social media. Because readers must fill in the missing part themselves, they often disagree on whether a text contains one. This paper asks how to detect and evaluate enthymemes when such disagreement is part of the task. On tweets about vaccination and immigration, each labeled by five annotators, F1 stalls around 0.67 for every fine-tuned model we train, regardless of data size, added features or label type. We therefore ask how these models should be evaluated instead, and propose metrics that compare predictions with the full distribution of annotator votes, together with a model that predicts this distribution directly and captures disagreement better at no cost in F1. We then ask what role discourse structure plays. Models learn to rely on explicit argumentative structure: they reliably recognize tweets without an enthymeme but rarely commit on tweets that contain one. Both the models and a discourse parser miss arguments whose links are left implicit and must be inferred by the reader. Finally, we discuss what these blind spots mean for content moderation systems trained on majority labels.
\end{abstract}

\section{Introduction}
\label{sec:introduction}

This paper investigates the detection of arguments with missing premises or conclusions, known as enthymemes \citep{WALTON2008361,raz}, in short texts. The study is motivated by the need to detect enthymemes in social media, where they are a frequent means of persuasion: by leaving part of the argument unstated, they lower the critical attention paid to its implicit content, which eases the acceptance of fallacies and false information and facilitates deceptive argumentation \citep{jbp:/content/journals/10.1075/jaic.00009.lom}. 

For example, in the tweet \textit{``Life isn't a movie and vaccines aren't created in 24hrs. The covid vaccine doesn't work and never has.''}, the missing component is a premise:

\begin{center}
\begin{tabularx}{0.85\linewidth}{@{}l X@{}}
$P_1$: & Vaccines aren't created in 24 hours \\[4pt]
$P_2$: & \textcolor{blue}{The COVID vaccine was created in 24 hours \textit{(implicit premise)}} \\[4pt]
\multicolumn{2}{@{}l@{}}{\hrulefill} \\[4pt]
$C$: & The COVID vaccine doesn't work and never has \\
\end{tabularx}
\end{center}

\noindent Enthymemes cannot however be approached as a standard detection task evaluated, since their presence is not determined by the explicit text alone. Detecting one depends on how a reader reconstructs the missing component, so that accounting for variation in human judgment is part of the task. We therefore take a perspectivist approach that models not only whether an enthymeme is detected, but also how consistently it is recognized across readers.

Accordingly, one of the challenges we take up in this paper is how to evaluate enthymeme detection under human label variation and disagreement. Once the multiple annotations are kept, the majority-vote verdict is no longer the only object to be predicted, and several properties of the vote distribution become relevant targets \citep{plank-2022-problem,uma2021learning}. We address this challenge in three steps. First, we bring together metrics from work on learning and evaluating with disagreement \citep{baan-etal-2022-stop,rizzi-etal-2024-soft,leonardelli-etal-2025-lewidi} and add a per-vote metric conditioned on annotator behavior (Section~\ref{sec:evaluation}, and Section~\ref{sec:pervote} for the per-vote metric). Second, we introduce betabin, a model that predicts a distribution over the vote count, thereby separating the leaning of a tweet from the expected division among its readers (Section~\ref{sec:betabin}). Then follows an error analysis organized by the variation metrics (Section~\ref{sec:analysis}). Its goal is to make the variation visible and to explain it. We study where the betabin model gains over the others, and what the tweets it gains on have in common in terms of discourse structure. 

Running alongside the above objective of measuring variation is a second complementary
thread that concerns discourse structure. Here we use discourse structure not only as a
source of predictive information, but also as a way to understand what makes an enthymeme
easier or harder to detect and where human judgments diverge. We first encode the RST tree
of each tweet as a set of features and inject them into the detection model, following the
discourse-enhanced architecture of \citet{chernyavskiy-etal-2024-unleashing}, which tests
whether the structure carries information that the text alone does not (Sections~\ref{sec:features} and~\ref{sec:injection}). We then reuse the
same features as descriptive dimensions in the success and error analysis announced above,
to characterize the tweets on which the models succeed, fail, and place near-threshold
probabilities (Section~\ref{sec:analysis}). To further examine how the model interacts with discourse structure, we
probe the encoder before and after fine-tuning and ask which discourse configurations it
represents and which the detection task strengthens (Section~\ref{sec:probing}). The three steps together let
us examine discourse structure from both sides. On the model side, they identify the
discourse profiles a system can rely on to detect an enthymeme under label variation
and those it fails to read, which lead in turn to failures in capturing the phenomenon. On
the reader side, the same dimensions describe the tweets on which annotators divide, which
gives a discourse-based account of interpretive variation and disagreement.

We finish by discussing what the variation metrics make visible, and
what this implies for pipelines that classify and moderate social media content on the
basis of majority labels (Section~\ref{sec:discussion}). We first give F1 its due and set out why our data leaves little
room for improving on it (Section~\ref{sec:f1_case}). We then describe three distortions of the discourse ecosystem
that a verdict-based evaluation leaves unmeasured, and how a distributional output
addresses each (Section~\ref{sec:ecosystem}).

The main contributions of the paper are as follows:

\begin{itemize}
  \item An evaluation of enthymeme detection under human label variation. We bring together
  metrics from work on learning with disagreement, organized by the question each answers
  about the vote distribution, and add a per-vote metric that tests whether a model's
  errors fall on the votes that are unusual for the annotator who cast them.

  \item Betabin, Section~\ref{sec:betabin}, a model that predicts a distribution over the vote count rather than a label. Unlike soft-label training, which fits a single vote fraction, it predicts a
  leaning and a confidence in that leaning. It matches the text-only baseline on macro-F1,
  improves on every measure of vote distribution shape, and its output is interpretable in
  vote units.

  \item Evidence that F1 is bounded at around 0.67 regardless of data volume, feature
  injection, or label regime, from a systematic comparison of feature-only, classical and
  transformer systems with learning curves for each. The properties of the vote
  distribution are learnable where the verdict is not.

  \item A discourse-structural account of where detection succeeds and fails. Injected RST
  features are redundant, but the same dimensions used descriptively show that the models
  rely on explicit argumentative structure as a proxy for the label, that the certain
  absence of an enthymeme is learned while its certain presence is not, and that a reported
  speech and evaluation configuration marks the tweets on which readers divide.

  \item Probing evidence that fine-tuning promotes explicit stance and attribution
  configurations but not implicit marking, and sheds trope content, which accounts for the
  redundancy of the injected RST features and for the failure on unmarked junctions.
\end{itemize}

\section{Related Work}
\label{sec:related_work}

\subsection{Enthymeme Detection}
\label{sec:rw_enthymemes}

The term enthymeme first appears in Aristotle's \textit{Prior Analytics}, where it names an incomplete syllogism, an argument with a proposition left unstated \citep{WALTON2008361}. This deductive reading, the enthymeme as a truncated syllogism whose missing part can be supplied to restore validity, persisted into modern logic \citep{Hitchcock1985}, and has since given way to a rhetorical and pragmatic conception. On this account the enthymeme is a form of defeasible reasoning that holds unless there is evidence to the contrary, and its missing component is omitted because the audience is expected to supply it, which is the source of its persuasive force \citep{walton2008argumentation,raz}. A contestable proposition introduced implicitly can therefore enter discourse without the scrutiny an explicit assertion would attract \citep{reboul2011relevance,jbp:/content/journals/10.1075/jaic.00009.lom}.

Detecting whether a text is enthymematic has received less attention than recovering the implicit component. Recovery itself has been pursued through logic-based decoding \citep{hunter2022understanding}, LLM-assisted logical reconstruction \citep{feng-hunter-2026-making}, and neural generation drawing on commonsense knowledge \citep{chakrabarty-etal-2021-implicit,stahl-etal-2023-mind}. For enthymemes, detection has been approached with feature-based classifiers \citep{rajendran-etal-2016-contextual} and fine-tuned neural models \citep{stahl-etal-2023-mind}, though on data constructed differently from ours: their texts are made enthymematic by deleting a component, so every item contains one by design, whereas our tweets occur naturally and may contain none. These lines of work do not, however, study detection under label variation, for lack of a resource that keeps more than one annotation per item.

\subsection{Discourse Structure and Implicit Argumentation}
\label{sec:rw_discourse}

Efforts to detect implicit argumentative content have mostly drawn on external knowledge rather than on discourse structure. \citet{chakrabarty-etal-2021-implicit} augment a generation model with commonsense knowledge to recover implicit premises, and \citet{Becker2024} reconstructs implicit knowledge from commonsense resources, while noting that surface discourse markers can mislead a system as often as they guide it. The role of discourse structure itself in signalling implicit argumentation has received little attention.

The main exception is the work of \citet{Green2010-GREROA}. She proposes ArgRST, a hybrid representation that encodes argument components (data, claim, warrant, backing) within an RST tree, mapping them onto relations such as \textsc{evidence}, \textsc{antithesis} and \textsc{concession}. On a corpus of genetic-counseling letters, she uses this representation to characterize the discourse contexts that license the omission of an argument element. In ArgRST, omitted argument components are reconstructed as implicit propositions within the discourse tree. Her analyses therefore provide a precedent for using discourse structure to characterize the contexts in which implicit argumentation occurs.

To our knowledge, no study has further examined whether discourse-structural relations can be used to detect the presence of an implicit premise or conclusion in short text, still less how that signal relates to the variation among readers who judge it.

\subsection{Perspectivist NLP and Variation}

NLP evaluation has traditionally assumed that each instance has a single correct label. Disagreement between annotators is therefore commonly treated as annotation noise to be resolved through aggregation. \citet{aroyo2015truth} challenged this view by showing that disagreement can reflect differences in how people interpret and annotate the same instance. This perspective has since developed into perspectivist NLP, which treats variation between human judgments as potentially informative for data collection, modeling, and evaluation \citep{basile-etal-2021-need,plank-2022-problem,cabitza2023perspectivist}.

This shift has motivated methods that retain disagreement in both learning and evaluation. \citet{uma2021learning} survey approaches to learning from disagreement, including methods that model individual annotators and methods that represent multiple judgments directly. \citet{gordon2021disagreement} show that aggregated labels can obscure the distinction between model error and disagreement between human judgments, and propose disagreement deconvolution to separate these sources of variation. \citet{gordon2022jury} extend this perspective to model training with Jury Learning, which incorporates individual judgments rather than reducing them to a single label.

More recent work has focused on metrics for evaluating predictions against distributions of human judgments. \citet{baan-etal-2022-stop} argue that calibration should be evaluated against human judgment distributions when annotators disagree. \citet{rizzi-etal-2024-soft} compare soft metrics for evaluation under disagreement and show that different metrics capture different properties of the annotation distribution. The LeWiDi shared tasks have further developed evaluation of systems that learn from disagreement \citep{leonardelli-etal-2025-lewidi}.

Enthymeme detection provides a relevant setting for this perspective. Identifying an implicit premise or conclusion requires semantic interpretation of the text and its possible argument structure. Different readers can therefore arrive at different judgments about whether an implicit component is present. In this setting, variation between judgments provides information about how the text can be interpreted.

\section{Task, Dataset, and Annotation}
\label{sec:task_dataset}

This paper draws throughout on the resource of \citet{pastor2026enthymeme}, which
supplies the formal definition of the task, the annotation guidelines, and the annotated
tweets, including the five votes per item on which the evaluation of Section~\ref{sec:evaluation} rests. We restate below the parts of that framework the rest of the paper depends on.

\subsection{Enthymemes}
\label{sec:task_definition}

An enthymeme is an argument with a missing component. Given premises $p_1$, $p_2$ and a conclusion $c$, one of the three is absent from the text. We follow the definition of \citet{pastor2026enthymeme}, which grounds the reconstruction of the missing component in Walton's argumentation schemes \citep{walton2008argumentation}. The tweet below, from \citet{burke_vlah_2023}, leaves its second premise unstated. The reconstruction is licensed by the scheme \textit{Argument from Negative Consequences}. 

\begin{quote}
\textit{``I don't wanna get a COVID vaccine because some days I would like to have babies''}
\end{quote}

\begin{center}
\begin{tabularx}{0.9\linewidth}{@{}l X@{}}
$P_1$: & I want to have babies \\[3pt]
$P_2$: & \textcolor{blue}{The vaccine causes infertility \textit{(implicit)}} \\[3pt]
\multicolumn{2}{@{}l@{}}{\hrulefill} \\[3pt]
$C$: & I don't want to get the COVID vaccine \\
\end{tabularx}
\end{center}

Because reconstructing an implicit component is otherwise open to subjective or distorted readings, the framework of \citet{pastor2026enthymeme} constrains it: the inferential license of the argumentation scheme is reified as a propositional conditional and added to the premise set, so that the reconstructed premises classically entail the conclusion. Entailment serves as a completeness check on the reconstruction. 

Furthermore, not every unstated proposition counts as a missing component. The annotation guidelines of \citet{pastor2026enthymeme} annotate an implicit component only when it is controversial. In these guidelines, a controversial component expresses a stance on a state of affairs, that is, a position on values, beliefs or attitudes, of the kind that can influence the convictions, choices and behavior of other people. An implicit component that carries no stance, committing the author to nothing, is out of scope and annotated as \textsc{none}. The dataset restricts this criterion to two topics, vaccination and immigration. 

\subsection{Task}
\label{sec:task}

Each annotation assigns one of three labels to a tweet, \textit{none}, \textit{implicit premise} or \textit{implicit conclusion}, and records the propositional decomposition in the format above together with the argumentation scheme. In this work we merge the two implicit labels into a single \textit{enthymeme} label and treat detection as a binary task: enthymeme against none. Each tweet carries the votes of five annotators. As we will see below, we keep these votes rather than reducing them to a majority label, and the evaluation of Section~\ref{sec:evaluation} is built on them.

\subsection{Dataset and Annotation}
\label{sec:dataset}

The dataset of \citet{pastor2026enthymeme} contains 1,482 English tweets drawn from the trope dataset of \citet{flaccavento-etal-2025-automated}, which was collected with keyword queries on vaccines and immigration between late 2019 and late 2023. The tweets were sampled to balance trope against no-trope labels. Vaccination accounts for 1,169 tweets (78.9\%) and immigration for 313 (21.1\%). Each tweet carries five votes, which are used throughout this work. By majority vote, \textit{none} accounts for 980 tweets (66.1\%), \textit{implicit premise} for 439 (29.6\%) and \textit{implicit conclusion} for 63 (4.3\%).

\subsection{Trope and RST Annotations}
\label{sec:tropes}
\label{sec:rst_annotations}

This paper works mainly with parser-generated RST features, both for the experiments and as the basis for the later discourse analysis. Alongside them we use the gold trope labels of the dataset the enthymeme resource was built on, to compare two possible linguistic means of signalling an enthymeme: discourse structure, which captures the formal organization of a tweet, and online tropes, which invoke recurring, culturally situated figures readers recognize.

\textbf{Tropes} have been defined in media studies as a storytelling device or shortcut that assumes the audience recognizes the situation \citep{gala-etal-2020-analyzing}. By online trope we mean such a device used in online discussion. Online tropes refer to human situations rather than plots, are often implied rather than stated, and are readily understood by readers. Examples include recurring situations such as distrust experts, hidden agenda, and natural is better. Every tweet carries the gold trope labels of \citet{flaccavento-etal-2025-automated}, which cover nine trope categories. Tropes describe the content of a tweet, not its form.

\textbf{Rhetorical Structure Theory} (RST; \citeauthor{mann1988rhetorical} \citeyear{mann1988rhetorical}) describes the coherence of a text as a tree. The text is segmented into elementary discourse units (EDUs), clause-like spans, and EDUs are linked by coherence relations such as \textsc{elaboration}, \textsc{contrast}, \textsc{cause} and \textsc{temporal}. In each relation one unit, the nucleus, carries the essential content, and the other, the satellite, adds to it. These relations combine recursively, so that a whole tweet is represented as a single tree, of the kind shown in Appendix~\ref{sec:appendix_rst_trees}. Some relations, such as \textsc{joint} and \textsc{same-unit}, are multinuclear. RST trees for all tweets were produced automatically with the IsaNLP RST parser \citep{chistova-2024-bilingual}, using its model trained on the RST Discourse Treebank, with the harmonized set of 18 relation labels of \citet{braud-etal-2017-cross}. Section~\ref{sec:features} describes how the trees and the trope labels are encoded as features.

\section{Modeling Enthymeme Detection}
\label{sec:modeling}

We begin with the standard methodology: enthymeme detection framed as binary supervised classification — enthymeme vs. no enthymeme — evaluated against gold labels.\footnote{We do not distinguish implicit premises from implicit conclusions: both instantiate the same phenomenon, an unstated argumentative component, and our questions concern how discourse structure signals its presence, not which component is missing.} Our objective is to find the best performing detection model under gold-label F1, and to measure whether the explicit use of external features such as tropes and RST discourse features helps.

\subsection{RST and Trope Feature Representations}
\label{sec:features}

Our RST features follow, as a starting point, the encoding proposed by \citeauthor{chernyavskiy-etal-2024-unleashing} (\citeyear{chernyavskiy-etal-2024-unleashing}) for propaganda detection, which represents each text through per-EDU nuclearity and relation information together with global tree statistics such as the number of EDUs, tree depth, and relation counts. However, a feature-only diagnostic revealed that on short tweets these count-based features largely reduce to proxies for text length: a logistic regression on the features alone barely outperforms one trained on length alone, and once the length-correlated component is removed from the features, performance drops below the length baseline. 

We therefore constructed a compact 24-dimensional encoding built exclusively from indicators and ratios, making it insensitive to text length.  These more focused features emphasize discourse structure, that is, which relations organize the tweet, how they are configured, and how the tree is shaped, rather than how much material it contains: they record the presence and placement of the core relations, the proportions in which relations occur, nuclearity and satellite ordering, coordination patterns such as root and embedded \textsc{joint}, and global shape ratios over the nuclear spine and branching. Table~\ref{tab:v9_features} in Appendix~\ref{sec:appendix} lists each feature together with the motivation for its selection.

The tropes are encoded as a nine-dimensional binary vector of gold trope annotations covering all 1,482 tweets. Unlike the RST features, tropes require no structural transformation: they operate at the level of content rather than form. 

Features-only diagnostics place tropes alone at 0.538 macro-F1 and the combined 33-dimensional RST and tropes set at 0.608, so both feature families are individually predictive before any injection into a neural model.

\subsection{Classical Approaches}
\label{sec:classical}

Before turning to fine-tuned transformers, we experiment with classical approaches in order to establish baselines and to investigate the role played by RST and trope features at this level. We train a balanced Logistic Regression and a linear SVM \cite{cortes1995support}, implemented via Scikit-learn (\citeauthor{pedregosa11a}, \citeyear{pedregosa11a}; \citeauthor{buitinck:hal-00856511}, \citeyear{buitinck:hal-00856511}), on combinations of three feature blocks: a TF-IDF text representation \cite{salton1988term} with word 1-2-grams and character 3-5-grams, the 24-dimensional RST features, and the trope flags. All models are evaluated out-of-fold on the same 5-fold partition later used for the neural experiments, and on the held-out test set of 148 tweets, with vectorizers and scalers fit on the training data only. Logistic Regression is the stronger model in every combination, so we report its scores in Section~\ref{sec:results}.

\subsection{Fine-Tuned Transformers with Hard vs. Soft Labels}
\label{sec:transformers}

We use DeBERTa-v3-base \cite{he2023debertav3} as our detection model. Fine-tuned DeBERTa has proven effective for detecting implicit content in short texts in a setting close to ours \cite{stahl-etal-2023-mind}, and as an open encoder it lets us intervene on the model itself: inject external features into its architecture, and later probe its internal representations. We fine-tune with a maximum length of 128 tokens, an effective batch size of 32, a learning rate of 1e-5 with layer-wise decay, and up to 10 epochs with early stopping on validation loss.

Each tweet in our dataset carries the votes of five annotators. This gives us access not only to a majority-vote label, but also to the label variation behind it, which is what is needed for a semantic interpretation task like enthymeme detection, where the judgment genuinely varies across readers and a majority vote erases the interpretation diversity behind the label. 

We therefore train under two regimes. In the hard regime, the target is the majority vote and the loss is a class-weighted cross-entropy. In the soft regime, the target is the vote fraction, treated as a two-way distribution; the loss is a soft focal cross-entropy \cite{lin2017focal}, which down-weights confidently wrong predictions on contested items to the degree the annotators themselves were split. Everything else is held identical, so any difference in behavior is attributable to the choice of target. Results are reported in Section~\ref{sec:results}.

\subsection{Feature Injection Architectures}
\label{sec:injection}

A first way to inject features is to concatenate them with the text representation, following \citeauthor{chernyavskiy-etal-2024-unleashing} (\citeyear{chernyavskiy-etal-2024-unleashing}), whose discourse features improved propaganda detection on news paragraphs. We experimented with their approach, both their original encoding and the concatenation-based injection, but it performed poorly in our setting of much shorter texts. 

Concatenation makes the features available to the classifier for every prediction. Our feature-only diagnostics show that the features predict the label well above chance but well below the text representation, so on most tweets they add little to what the text already provides. We therefore use an architecture that can consult the features selectively or ignore them entirely with a learned gate weights the feature representation before it is added residually to the text representation.\footnote{Full architectural details are available in our code release: \url{https://github.com/metabolean5/enthymeme-variation}.} With the gate closed, the model degrades to the baseline: the injection can only add information, never displace the text. To stabilize the fused system on our small dataset, we freeze the parameters of the lower half of the encoder layers, which helps only the gated models, not the baseline.
 
This architecture is run with three feature sets: the 24-dimensional RST features, the 9 trope flags, and their concatenation. 

\subsection{Betabin: Modeling the Votes}
\label{sec:betabin}

The architectures above inject information at the input. Lastly, we suggest injecting nothing and instead changing what is modeled, namely, the vote distribution itself. If disagreement is part of the phenomenon, then the natural training target is not the verdict, nor even the vote fraction, but the annotation process: five readers casting votes. A model trained this way must learn not only which way a tweet leans but how divided its readers will be, a property we will evaluate directly with the disagreement metrics of Section~\ref{sec:evaluation}.

Beta distributions over annotator response probabilities go back to classical annotation modeling \cite{dawid1979maximum,paun-etal-2018-comparing}; we introduce this objective for disagreement modeling in our setting. The head predicts, for each tweet, the two parameters of a Beta distribution over the probability that a reader votes enthymeme, and training maximizes the likelihood of the observed vote count (e.g., 3 of 5) under it.\footnote{The two parameters are kept positive with a softplus, and the loss is this likelihood alone, with no other terms.} The prediction thus carries two degrees of freedom instead of one: a leaning, and a confidence in that leaning, so the model can distinguish ``I expect readers to split on this tweet'' from ``I do not know this tweet'', two beliefs a single probability collapses into the same hedged output. At evaluation, the predicted mean drops into the same thresholding pipeline as every other system, so hard-label F1 remains directly comparable.

\subsection{Results on Hard-Label F1}
\label{sec:results}

\begin{table}[!tbp]
\centering
\caption{Test-set results (148 tweets; 97 none, 51 enthymeme). P = precision, R = recall. The \textbf{None} and \textbf{Enthymeme} columns give per-class scores, treating none and enthymeme in turn as the positive class. All systems were developed on the 1,334-tweet train+dev pool, with selection based on out-of-fold cross-validation scores, and evaluated on the test set once. Shallow models (Logistic Regression) are single deterministic fits; transformers report mean $\pm$ standard deviation over 5 seeds, with per-class scores from the seed-averaged predictions, at a fixed 0.5 threshold unless noted. The hard-label row at 0.5 includes one near-degenerate seed, which raises its standard deviation to .06 against .01 to .03 for all other transformer rows; its dev-tuned row repairs the threshold per seed. Best score per column in bold.}
\label{tab:test_results}
\small
\setlength{\tabcolsep}{6pt}
\renewcommand{\arraystretch}{1.0}
\begin{tabular*}{\textwidth}{@{\extracolsep{\fill}} l cc ccc ccc @{}}
\toprule
& & & \multicolumn{3}{c}{\textbf{None}} & \multicolumn{3}{c}{\textbf{Enthymeme}} \\
\cmidrule(lr){4-6} \cmidrule(lr){7-9}
\textbf{System} & \textbf{Acc} & \textbf{Macro-F1} & \textbf{P} & \textbf{R} & \textbf{F1} & \textbf{P} & \textbf{R} & \textbf{F1} \\
\midrule
\multicolumn{9}{@{}l}{\textit{Features alone (LogReg)}} \\[-1pt]
RST (24d) & .608 & .607 & .831 & .505 & .628 & .461 & .804 & .586 \\
Tropes (9d) & .547 & .544 & .734 & .485 & .584 & .405 & .667 & .504 \\
RST + tropes & .588 & .582 & .765 & .536 & .630 & .438 & .686 & .534 \\
\midrule
\multicolumn{9}{@{}l}{\textit{Classical text (LogReg)}} \\[-1pt]
TF-IDF & .588 & .554 & .696 & .660 & .677 & .411 & .451 & .430 \\
TF-IDF + RST & .622 & .605 & .753 & .629 & .685 & .463 & .608 & .525 \\
TF-IDF + tropes & .608 & .580 & .719 & .660 & .688 & .441 & .510 & .473 \\
TF-IDF + RST + tropes & .635 & .624 & .779 & .619 & .690 & .479 & .667 & .557 \\
\midrule
\multicolumn{9}{@{}l}{\textit{Transformers (5 seeds, mean $\pm$ sd)}} \\[-1pt]
Text-only, soft & .685 {\tiny$\pm$.02} & .671 {\tiny$\pm$.02} & .816 & .674 & \textbf{.736} & \textbf{.534} & .706 & .606 \\
Text-only, hard & .630 {\tiny$\pm$.06} & .627 {\tiny$\pm$.06} & \textbf{.903} & .489 & .626 & .485 & \textbf{.898} & .628 \\
Text-only, hard (dev-tuned $t$) & .666 {\tiny$\pm$.02} & .662 {\tiny$\pm$.02} & .865 & .584 & .695 & .510 & .824 & \textbf{.629} \\
+ RST (gated, soft) & .673 {\tiny$\pm$.02} & .667 {\tiny$\pm$.01} & .855 & .608 & .707 & .519 & .796 & .626 \\
+ Tropes (gated, soft) & .674 {\tiny$\pm$.03} & .650 {\tiny$\pm$.02} & .783 & \textbf{.703} & \textbf{.736} & .530 & .620 & .563 \\
Betabin (vote-count head, soft) & \textbf{.686} {\tiny$\pm$.02} & \textbf{.675} {\tiny$\pm$.02} & .822 & .666 & \textbf{.736} & .533 & .725 & .615 \\
\bottomrule
\end{tabular*}
\end{table}

Table~\ref{tab:test_results} reports all systems on the held-out test set of 148 tweets. Every model was developed exclusively on the pool of 1,334 train and development tweets: model and representation choices were made against out-of-fold scores on a fixed 5-fold partition of this pool, so that no decision ever consulted the test data, and each configuration was evaluated on the test set only once. Transformer results are averaged over five random seeds affecting parameter initialization and training; the shallow models are deterministic single fits and therefore carry no seed variance.

\subsubsection{Results of Features Alone}
\label{sec:results_features}

We first evaluate performance using the features alone without ever adding any tweet text as input. Both feature families carry signal on their own: RST features alone reach 0.583 macro-F1 out-of-fold (0.607 on test) and trope flags 0.538 (0.544), against a chance level of 0.398.\footnote{Macro-F1 of the majority-class predictor.} Their profiles differ in an instructive way (Table~\ref{tab:test_results}): the RST features over-call the enthymeme class, reaching the highest enthymeme recall of any shallow system (0.804) at the cost of none recall (0.505).

\subsubsection{Results of Classical Approaches}
\label{sec:results_classical}

Adding features to TF-IDF improves the classical systems modestly: TF-IDF alone reaches 0.554 macro-F1 on test, RST lifts it to 0.605, and the full combination tops out at 0.624.\footnote{None of the gains is significant; the largest carries a paired bootstrap 95\% confidence interval \cite{koehn-2004-statistical} spanning zero. Logistic Regression outperformed the linear SVM in every combination, so we report it throughout.} The RST gain is largely stability: TF-IDF falls 5.6 points from cross-validation (0.610) to test, while the RST features are the only block that improves out of sample (0.583 to 0.607) and, added to TF-IDF, restore most of the lost transfer. The per-class profiles in Table~\ref{tab:test_results} show where the features act: they chiefly raise enthymeme recall (0.451 to 0.667 for the full combination), correcting TF-IDF's failure on the minority class. The classical ceiling nonetheless remains well below the fine-tuned transformers, raising the question of the next subsection: whether injecting these same features helps where they are no longer the model's only source of signal.

\subsubsection{Results of Transformer Approaches}
\label{sec:results_transformers}

The text-only soft-label baseline reaches 0.671 $\pm$ 0.02 macro-F1 on test, 4.7 points above the best classical system. Three comparisons against this baseline structure the results.

\noindent\textbf{Hard vs.\ soft labels.} Soft training beats hard training modestly but consistently: across all seed-paired comparisons the soft model gains around one point, and it never lost to its hard twin in any pairing. At a fixed 0.5 threshold the gap widens to 4.4 points (0.671 vs.\ 0.627), but most of this is calibration: hard training with class weights inflates the probabilities, so that at 0.5 the model over-calls the enthymeme class (recall 0.898, none recall 0.489), and per-seed threshold tuning on the development set is needed to recover 0.662. The soft model requires no such repair, as its tuned thresholds all fall within 0.49 to 0.52. Soft training thus self-calibrates the decision boundary that hard training must fix post hoc.

\noindent\textbf{Feature injection.} No injection beats the baseline. The gated RST system reaches 0.667 $\pm$ 0.01, the gated tropes system 0.650 $\pm$ 0.02, and every difference has a confidence interval spanning zero. The details of the null are themselves informative. The RST system is the only configuration whose test score exceeds its cross-validation score, and it shows the lowest seed variance of any system we trained: consistent with the shallow results, structure is a stable signal, only a redundant one. The tropes are the sharper case: gold annotations, injected as an oracle unavailable at real inference time, produce the best cross-validation score of any configuration  and then fail to transfer to the test set.\footnote{We evaluated seven pre-registered feature-injection configurations on the test set in total, including the encoding of \citeauthor{chernyavskiy-etal-2024-unleashing} (\citeyear{chernyavskiy-etal-2024-unleashing}) that motivated this line of work; none beat the text-only baseline.}

\noindent\textbf{Betabin.} The vote-modeling architecture reaches 0.675 $\pm$ 0.02, nominally the highest test score of all systems in Table~\ref{tab:test_results}, though seed-paired against the baseline the difference is noise: the correct reading is parity. It is nonetheless the only "non-baseline" configuration in all our test shots that does not lose to the text-only soft, and it does so while predicting a strictly richer object, a full distribution over vote counts rather than a label. On gold-label F1, that additional structure is invisible by construction. Indeed, a model of the verdict and a model of the votes score the same. Whether the votes were worth modeling is exactly what the disagreement metrics of Section~\ref{sec:evaluation} are built to see. These metrics compare the predicted probability with the five-vote distribution of each tweet rather than with its majority label, and score whether the model locates the contested tweets, how far its probabilities sit from the vote fractions, and whether they are calibrated in vote units.

\begin{figure}[!tbp]
    \centering
    \includegraphics[width=0.6\textwidth]{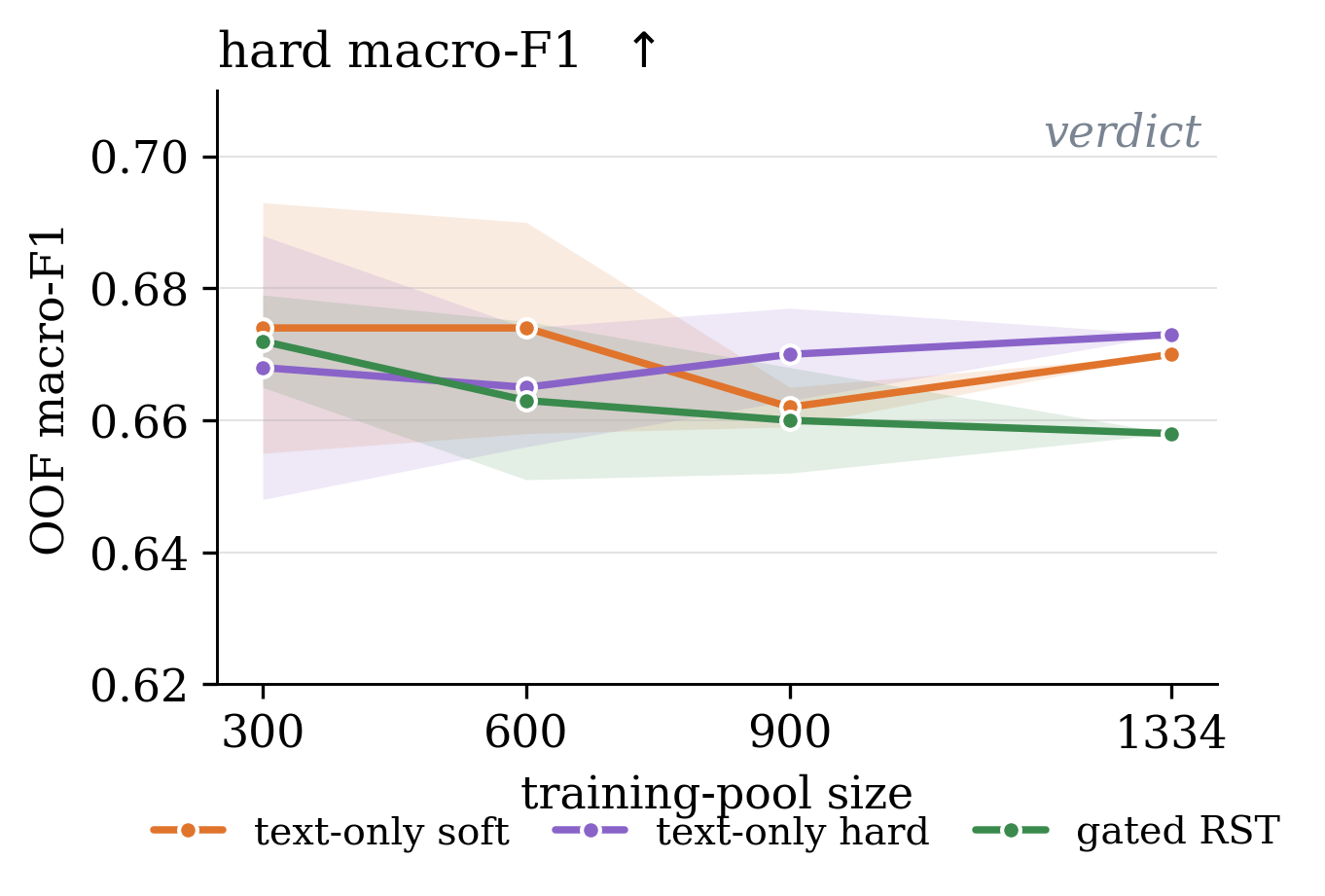}
    \caption{Out-of-fold (OOF) macro-F1 as a function of training-pool size (honest cross-validation, per-fold tuned thresholds). The arrow next to the metric name gives the direction of improvement ($\uparrow$: higher is better). Shaded bands give $\pm$1\,sd over repeated draws; the full-size points (1{,}334) carry no standard deviation, as they use the entire training pool and admit no repeated draws.}
    \label{fig:learning_curve}
\end{figure}

\subsection{Learning Curves and the Performance Ceiling}
\label{sec:learning_curves}

To understand what limits performance, we further trained three systems on stratified subsets of increasing size (300, 600, 900, and the full 1{,}334 examples), all on identical draws so the curves are directly comparable: the text-only model under both label regimes, and the gated RST system. Figure~\ref{fig:learning_curve} shows the resulting learning curves, and three observations emerge from them together with Table~\ref{tab:test_results}.

First, we observe that more data does not help. All three curves stay within a band of under two points from 300 to 1{,}334 examples; every step along the curve is smaller than the spread across draws at $n=300$, and which 300 tweets a model sees matters more than seeing four times as many. Second, the RST features are redundant rather than harmful. The gated system is the best-behaved model at $n=300$, where the feature branch steadies small-data training, and declines mildly as data grows: as the encoder sees more text, it appears to acquire on its own what the features carry, until the fusion pathway is pure overhead. The features lose to the baseline not because discourse structure is irrelevant but, we will argue, because it is already represented, a hypothesis we test directly in Section~\ref{sec:probing}. Third, with thresholds tuned, hard and soft training are indistinguishable on macro-F1 at every training size, even though they are trained toward different targets and, as we will see in the next section, learn measurably different things.

Macro-F1 thus remains around 0.67 regardless of data volume, feature engineering, or label regime; across all our transformer configurations, scores cluster between 0.65 and 0.69. One plausible reading is that this reflects a property of the task rather than of the models: enthymeme detection is an interpretation task, and our five annotators disagree on 37\% of items. Our data shows early signs in this direction: model errors concentrate on the contested items, and the models place their probabilities differently on contested and consensus tweets. Gold-label F1, which scores every prediction against a single collapsed verdict, cannot register any of this. In the next sections we turn to what lies behind the label: the shape of the disagreement the majority vote collapses, what makes items contested, and how to measure what the models have learned about this variation.

\section{Evaluation Under Annotator Disagreement}
\label{sec:evaluation}

Human label variation was long treated as noise to be minimized under the assumption of a single ground truth; in NLP evaluation literature, a growing body of work instead treats it as signal, reflecting genuine differences in interpretation \cite{plank-2022-problem,uma2021learning}, and calls for models to be evaluated on their ability to recognize this variation \cite{leonardelli-etal-2025-lewidi}. Enthymeme detection belongs squarely in this class and the five-way annotations present in the dataset makes the variation measurable.

This section proposes an evaluation for this setting. We first review why gold-label F1 falls short here, then assess existing disagreement-aware metrics and what they capture for enthymeme detection, and finally propose a metric targeting a distinction the existing ones miss: whether a model's errors fall on the votes that are unusual for the annotator who cast them, which are the votes most likely to reflect inconsistent annotation rather than genuine disagreement.

\subsection{Why Gold-Label F1 Falls Short}
\label{sec:f1_falls_short}

\begin{figure}[!tbp]
    \centering
    \includegraphics[width=0.72\textwidth]{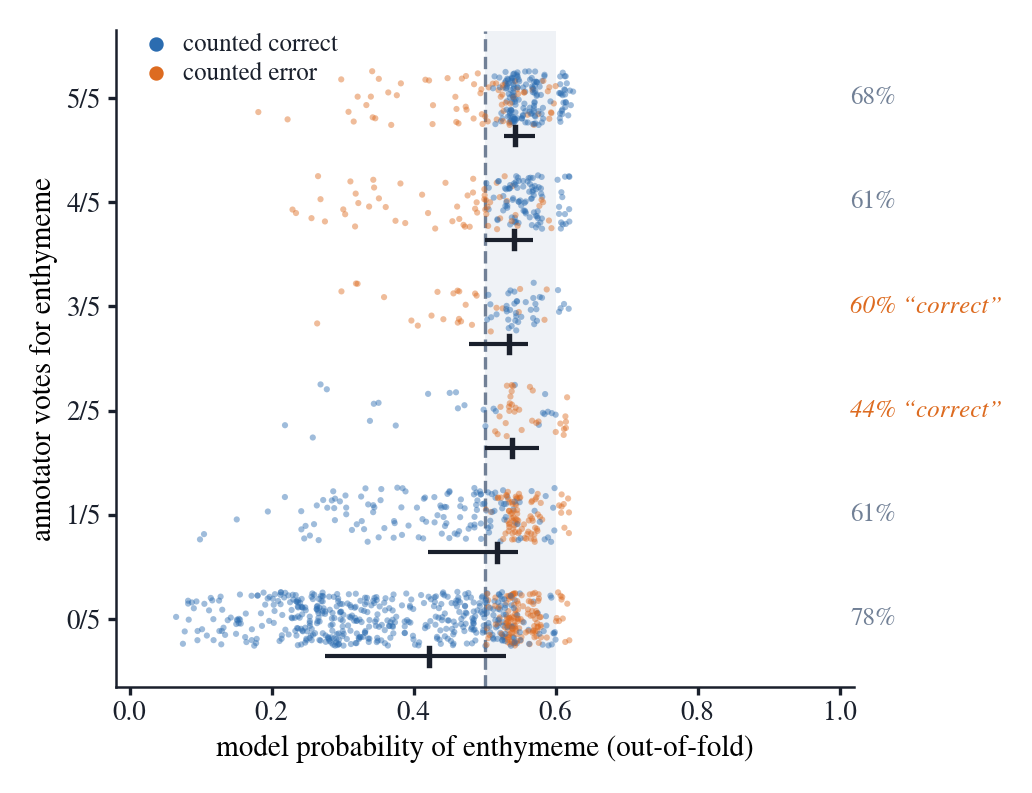}
    \caption{Out-of-fold probabilities of the text-only soft baseline by annotator vote band (points jittered vertically within each band; horizontal bars give the band mean). Blue points are counted correct by F1, orange as errors; the dashed line marks the 0.5 threshold. Only the 0/5 band spreads across the scale. The bands from 2/5 to 5/5 all sit between 0.5 and 0.62 and overlap, so contested tweets and unanimous enthymemes receive the same probabilities, a difference F1 does not register.}
    \label{fig:f1_blindspot}
\end{figure}

Enthymeme detection is a semantic interpretation task: identifying an implicit premise requires segmenting the text into propositions, recovering its argument structure, and drawing on commonsense, speech act perception, and background knowledge, processes that legitimately vary across readers. Analyzing disagreement on this dataset, \citet{pastor-vanarkel-2026-disagreement} show that a substantial part of it is genuine, grounded in valid interpretive variation (which reconstruction of a multi-proposition tweet is soundest, how a rhetorical question or an argument scheme is perceived) rather than annotator lapses. If disagreement is part of the phenomenon, an evaluation that first collapses five judgments into one verdict discards precisely the variation to be modeled: what should be measured is not only whether a model matches the majority, but whether it captures the shape of the disagreement.

The second reason is empirical, from the modeling side. Figure~\ref{fig:f1_blindspot} shows the out-of-fold probabilities of the text-only soft baseline by annotator vote band. The model has learned that the task is ambiguous: its probabilities all sit between 0.07 and 0.62, unanimous enthymemes receive barely more than 0.5, and the two contested bands are indistinguishable from each other. The 0.67 ceiling is obtained by cutting this narrow band at a tuned threshold. F1 only asks which side of that cut each item falls on; it cannot tell a confident model from one that hedges everywhere, and it cannot see whether consensus and contested items receive different probabilities. Under F1, we are blind to how the annotators are actually divided.

\subsection{Capturing Disagreement with Standard Soft Metrics}
\label{sec:metrics}

\citet{rizzi-etal-2024-soft} assess existing soft metrics and find them non-equivalent, and model rankings shift with the chosen metric \cite{sarumi-etal-2025-nlp}. These shifts are expected once one notes that the metrics answer different questions about the vote distribution; we therefore report a set of metrics organized by the question each answers.

\subsubsection{Existing Metrics and What They Measure}
\label{sec:existing_metrics}

First we can ask \textit{where} the disagreement is. Ambiguity Spearman computes the rank correlation \cite{spearman1904proof} between the model's uncertainty\footnote{For the transformer systems, each metric in this section is computed per seed and averaged over the five seeds.} and the vote entropy of each tweet, and thus measures whether the model locates the contested items where the annotators do.

Next we can look at \textit{depth} following  \citeauthor{leonardelli-etal-2025-lewidi} (\citeyear{leonardelli-etal-2025-lewidi}): the Manhattan distance between the predicted probability and the vote fraction measures the average misplacement on the vote scale. We also run soft cross entropy which penalizes the same misplacement logarithmically, so that confident errors weigh more. 

Lastly we have \textit{calibration}: soft-ECE adapts expected calibration error to the vote distribution, following the argument of \citet{baan-etal-2022-stop} that calibration under disagreement must be measured against human judgment distributions rather than a gold label; it checks that the predicted probability matches the votes it announces, so that a tweet scored 0.6 indeed received about three enthymeme votes out of five. We also add consensus separation, computed as the difference in mean predicted probability between the unanimous enthymemes and the unanimous nones. It complements soft-ECE in that calibration is averaged over all tweets and can remain acceptable when errors at the poles are diluted by the contested bands, whereas the group-mean difference isolates the unanimous tweets and collapses to zero for the model that hedges everything near the threshold.

One further remark justifies the composition of our choice of metrics. The \textit{depth} metrics have a weakness, in that a model predicting close to 0.5 everywhere scores well on the contested items, simply because the vote fractions there sit close to 0.5 as well. Such a model looks accurate on these items without knowing anything about them. The \textit{where} and \textit{calibration} metrics are included because they catch this behavior: a model that predicts 0.5 everywhere cannot rank contested items above consensus ones, and its probabilities do not match the votes on consensus items.

\subsubsection{Results Under Soft Metrics}
\label{sec:soft_metric_results}

Table~\ref{tab:soft_metrics} reports the best classical system and the four transformer systems. Under hard F1 the soft-label transformers are at parity (.667--.675, within one standard deviation), well above the classical system (.624), while the hard-label model scores .627 at the fixed 0.5 threshold (Section~\ref{sec:results_transformers}). Under every \textit{where}, \textit{depth} and \textit{calibration} metric the transformer ordering is strict: betabin, then the text-only baseline, then gated RST. The classical system, despite its lower F1, beats gated RST on most shape metrics; for the shape of the vote distribution, injecting RST features may thus be considered harmful to the transformer, plausibly for the same reason it was useless on F1: the information is redundant with what the encoder already represents, and the added pathway only perturbs the output distribution.

Betabin reduces Manhattan distance by 13\%, raises consensus separation from .188 to .286, raises ambiguity Spearman from .172 to .282, regenerates the exact vote split 1.6 times more often (23.8\% vs.\ 14.6\%), and is 6.7 times better calibrated (soft-ECE .017 vs.\ .114). Its output is interpretable in vote units: a prediction of 0.6 means three of five annotators. Modeling the votes rather than the verdict thus costs nothing on F1 and further improves every measure of vote distribution shape.

These metrics nonetheless share a limit. They score the fit between predicted and observed distributions, but they are silent on the source of a disagreement, which can reflect valid interpretive variation as well as inconsistent annotation under task complexity \cite{pastor-vanarkel-2026-disagreement}. The available vote annotations do not themselves distinguish those sources. Moreover, a model can match a flat distribution without accounting for the individual votes that compose it: whether a particular vote is routine for its annotator or unusual. The next subsection proposes a metric that evaluates individual votes against annotator behavior. It does not determine whether a reading is licensed by the text: regularity in an annotator's voting is only a proxy for a stable perspective, and an unusual vote is not necessarily a lapse. It instead tests whether the model's errors align with this behavioral regularity.

\subsection{An Annotator-Conditioned Per-Vote Metric}
\label{sec:pervote}

The metrics above compare distributions per tweet and never look at an individual vote. Here, a \textit{dissenting vote} is a vote that differs from the majority of the other four annotators; it is defined only where those four votes have a majority. Not all such votes are alike: some may express valid minority readings, whereas others may reflect inconsistent annotation \cite{pastor-vanarkel-2026-disagreement}. The available annotations do not label that distinction. We therefore turn to the one additional signal in the vote matrix: \textit{annotator behavior}. 

A dissenting vote from an annotator who regularly casts such votes is behaviorally expected; the same vote from an annotator who almost never casts one is behaviorally unusual. The metric below asks whether a model's surprise at individual votes tracks this reference. Importantly, the model is not given annotator identities and makes one tweet-level prediction for all five votes. The metric therefore does not show that it predicts any one annotator's judgment. Rather, it tests whether the errors of that shared prediction are larger on votes that are unusual for their respective annotators.

Indeed, the five annotators differ in their voting behavior: the annotator who most frequently casts dissenting votes disagrees with the majority of the other four annotators up to twelve times as often as the most conformist annotator, and this behavior is stable across contexts. A dissenting vote that is routine for one annotator can be behaviorally unusual for another.\footnote{Estimated over the
1,334-tweet pool: the probability of voting against a unanimous majority among the other four annotators
majority ranges from .014 (A4) to .165 (A3), a twelve-fold spread; against
a 3--1 majority among the other four annotators, from .083 (A4) to .232 (A5). One annotator (A3)
casts dissenting votes at nearly the same rate in both contexts (.165 vs.\
.163), indicating a stable independent standard rather than
context-sensitive hesitation; A3 also has the strongest enthymeme lean on
evenly split items (.640 vs.\ .442--.559 for the others).} The per-vote
evaluation we introduce here aims not to treat all dissenting votes as interchangeable: the
same lone dissenting vote can be expected from one annotator and surprising from
another, which motivates the annotator-conditioned reference $d$
introduced below.

The metric is a Spearman rank correlation between two quantities computed for every vote, both introduced below: the model's \textit{surprise} at the vote, $s_{ij}$, and the vote's \textit{deviation} given who cast it, $d_{ij}$, for the vote of annotator $j$ on tweet $i$. First, we compute the model's \textit{surprise} at a vote. Let $v_{ij} \in \{0,1\}$ be the vote of annotator $j$ on tweet $i$ (1 for enthymeme) and $p_i$ the model's predicted probability of enthymeme; the surprise is the probability mass placed against the vote,
\[
s_{ij} = \bigl| v_{ij} - p_i \bigr|.
\]

Second, the vote's \textit{deviation} $d_{ij}$: how unexpected this vote is, given who cast it. For each annotator, we build a behavioral profile: how often they cast each vote when the other four annotators lean a given way (unanimously, three to one, or evenly). The deviation of a vote is then the improbability of that vote under its annotator's profile,
\[
d_{ij} = 1 - P_j\bigl( v_{ij} \mid \text{other-vote context} \bigr),
\]
where $P_j$ is annotator $j$'s estimated propensity to cast vote $v_{ij}$ in the given context of the other four annotators, computed over all tweets except tweet $i$. A dissenting vote from an annotator who regularly casts one is expected and receives a low deviation; the same vote from an annotator who rarely casts one receives a high deviation. The profiles never see the vote being scored and consult no majority label.

Third, the summary: the Spearman rank correlation $\rho = \mathrm{Spearman}\bigl( \{s_{ij}\}, \{d_{ij}\} \bigr)$ over all votes, which asks whether the model's surprise ranks votes the way the behavioral reference does. A rank correlation is used so that the compressed probability scales observed above cannot distort the result. Algorithm~\ref{alg:pervote} summarizes the full computation. We discuss what the metric reveals about our models in the next section.

\begin{algorithm}[t]
\caption{Annotator-conditioned per-vote metric}
\label{alg:pervote}
\begin{algorithmic}[1]
\Require model probabilities $p_i$, votes $v_{ij}$ for tweets $i$, annotators $j$
\ForAll{votes $(i,j)$}
    \State $s_{ij} \gets |v_{ij} - p_i|$ \Comment{step 1: model surprise}
\EndFor
\ForAll{annotators $j$, other-vote contexts $c \in \{$4--0, 3--1, 2--2$\}$}
    \State $P_j(1 \mid c) \gets$ share of enthymeme votes $j$ cast when the other four votes form context $c$
    \Statex \hspace{\algorithmicindent}\hspace{\algorithmicindent} (with Laplace smoothing; for each scored vote, recomputed leaving its tweet out)
\EndFor
\ForAll{votes $(i,j)$}
    \State $c \gets$ context formed by the other four votes on tweet $i$
    \State $d_{ij} \gets 1 - P_j(v_{ij} \mid c)$ \Comment{step 2: deviation given who cast it}
\EndFor
\State $\rho \gets \mathrm{Spearman}\bigl(\{s_{ij}\}, \{d_{ij}\}\bigr)$ over all votes \Comment{step 3: summary}
\State \Return $\rho$
\end{algorithmic}
\end{algorithm}

\begin{table}[!tbp]
\centering
\caption{Evaluation of the best classical system (TF-IDF + RST + tropes, Logistic Regression with probability outputs) and the transformer systems, grouped by the question each metric answers about the vote distribution. Test set (148 tweets; 5 seeds for transformers, single deterministic fit for the classical system), except soft-ECE and the annotator-conditioned per-vote metric (\textit{consistency}, ours, defined in Section~\ref{sec:pervote}), computed out-of-fold (1,334 tweets). Arrows give the direction of improvement; best score per row in bold.}
\label{tab:soft_metrics}
\scriptsize
\setlength{\tabcolsep}{2pt}
\begin{tabular*}{\textwidth}{@{\extracolsep{\fill}} l l c ccccc @{}}
\toprule
& & & \textbf{Best} & \textbf{Gated} & \textbf{Text-only} & \textbf{Text-only} & \\
\textbf{Metric} & \textbf{Question} & & \textbf{classical} & \textbf{RST} & \textbf{hard} & \textbf{soft} & \textbf{Betabin} \\
\midrule
Hard macro-F1 & majority verdict & $\uparrow$ & .624 & .667 & .627 & .671 & \textbf{.675} \\
\midrule
Ambiguity Spearman & \textit{where} & $\uparrow$ & .143 & .115 & .058 & .172 & \textbf{.282} \\
\midrule
Manhattan distance & \textit{depth} & $\downarrow$ & .331 & .346 & .339 & .323 & \textbf{.282} \\
Soft cross-entropy & \textit{depth} & $\downarrow$ & .656 & .627 & .674 & .597 & \textbf{.584} \\
\midrule
Soft-ECE & \textit{calibration} & $\downarrow$ & .097 & .116 & .083 & .114 & \textbf{.017} \\
Consensus separation & \textit{calibration} & $\uparrow$ & .198 & .138 & \textbf{.313} & .188 & .286 \\
\midrule
Per-vote $\rho(s,d)$ (ours) & \textit{consistency} & $\uparrow$ & .181 & .232 & .112 & .205 & \textbf{.239} \\
\bottomrule
\end{tabular*}
\end{table}

\begin{figure}[!tbp]
    \centering
    \includegraphics[width=\textwidth]{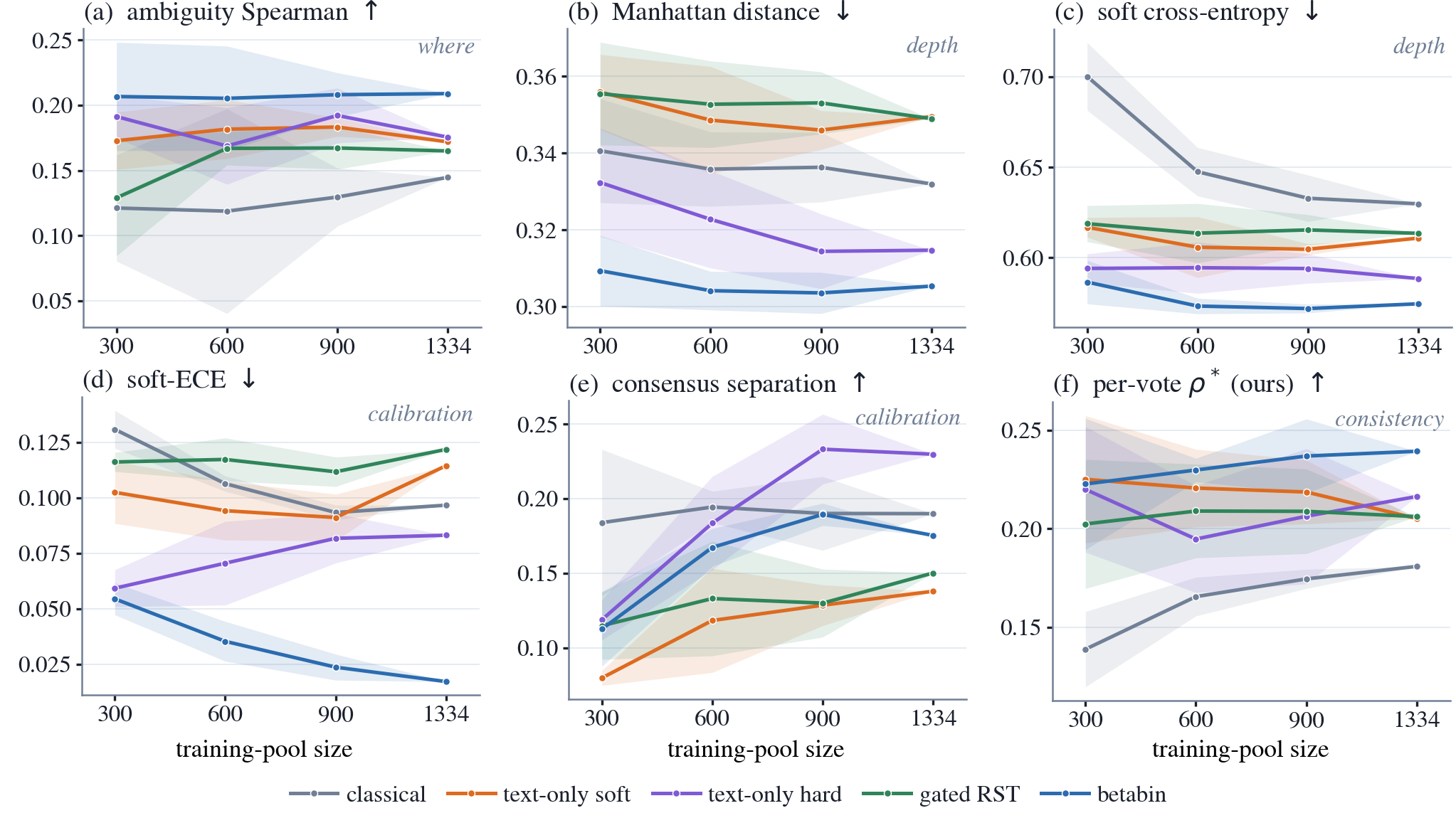}
    \caption{Soft metrics vs.\ training-pool size (out-of-fold; mean $\pm$1\,sd over repeated draws; arrows = direction of improvement). The full-size points (1{,}334) carry no standard deviation, as they use the entire training pool and admit no repeated draws.}
    \label{fig:soft_metrics_curve}
\end{figure}

\subsubsection{Results under the Annotator-Conditioned Per-Vote Metric}
\label{sec:pervote_results}

Figure~\ref{fig:soft_metrics_curve} (panel f) shows the metric across training-pool sizes, and the out-of-fold curves reproduce the ordering of Table~\ref{tab:soft_metrics}. Betabin improves with data and reaches the highest value at full size (.239 in Table~\ref{tab:soft_metrics}), while the classical system, despite improving steadily, remains below the soft-label transformers (.181). Gated RST inverts the distributional ordering: weakest of the transformers on the shape metrics, it is second on the per-vote metric (.232 vs.\ .205 for the text-only soft model), suggesting that the injected discourse features, which did not help for the verdict and for the distribution shape, carry some signal about which individual votes deviate.

On consensus tweets, our per-vote consistency metric behaves like a sharpened form of the depth and calibration measures, but with a different reference: rather than comparing the prediction with the vote fraction, it evaluates each vote against the behavioral profile of the annotator who cast it. Thus, a routine dissenting vote is discounted, and a model that commits against it is rewarded rather than penalized. As we will see in the error analysis, the tweets on which this discounting decides the ordering concentrate in one specific discourse profile.

\subsection{Summary and Scope of the Variation Metric Results}
\label{sec:metric_summary}

Figure~\ref{fig:soft_metrics_curve} shows the metrics across training-pool sizes. Across panels the pattern is consistent, in that hard F1 is flat from 300 examples on, while the distribution-level metrics change with data and the systems diverge under them. Betabin's calibration improves continuously (panel d), the hard model's consensus separation increases while its ambiguity Spearman remains low, and the gated RST system is flat or declining on nearly every panel. With more data, the models thus improve not on the verdict but on properties of the vote distribution, and the training regimes differ in which properties improve.

We should however reflect on how far these numbers take us. Each metric answers the question it was built for, whether the model locates the contested tweets, how far its probabilities are from the vote fractions, whether its outputs are calibrated in vote units, and the curves show that these properties are learnable where the verdict is not. What happens statistically behind a given improvement, why one regime improves calibration while another improves separation, is more than we can establish here, and this is not a statistics paper; the aggregate curves also do not show on which tweets the differences arise. What we can do is give these numbers a linguistic interpretation. The next section examines which tweets the models succeed on, fail on, and hedge on, and describes them by their discourse profiles.

\section{Discourse-Structure-Oriented Success and Error Analysis}
\label{sec:analysis}

Given that one of the goals of this paper is to understand the role discourse structure plays in enthymeme detection, and to ground the disagreements of our annotators in discourse-oriented terms, we proceed to an error and success analysis using the 24 RST feature dimensions\footnote{From this point on we speak of \textit{dimensions} rather than \textit{features}, reserving the latter for the quantities injected into a model and manipulated for prediction, as in Section~\ref{sec:results}. The same 24 quantities are used here descriptively, to characterize the tweets on which models succeed or fail, not as model inputs.} defined in Section~\ref{sec:features}, on the out-of-fold sample. Since the best transformer systems are indistinguishable on F1, the comparison operates entirely on the variation metrics for the success analysis. 

For the success analysis, we single out the tweets on which betabin scores consistently better than the second-best system, the text-only soft model, on each metric family in turn (\textit{where}, \textit{depth}, \textit{calibration}, \textit{consistency}), and ask whether discourse structure can explain these gains. For the error analysis, we turn to the tweets on which betabin fails across all metrics, and ask which forms of variation the model cannot capture, and what their discourse profile is.

\subsection{Success Cases on Consensus Tweets}
\label{sec:easy_cases}

We first ask what the transformer systems find easy under F1, and profile these easy cases with the RST dimensions of Section~\ref{sec:features} (Table~\ref{tab:v9_features}, Appendix~\ref{sec:appendix}). A tweet counts as \textit{easy} if all four transformer systems of Table~\ref{tab:soft_metrics} classify it correctly out-of-fold.\footnote{Correctness is computed on seed-averaged probabilities at per-fold tuned thresholds. Binary dimensions are compared by rate (Fisher's exact test), continuous ones by Cliff's $\delta$, following the recommendations of \citet{dror-etal-2018-hitchhikers}.} In what follows, ``easy enthymeme'' is shorthand for an easy tweet labeled as containing an enthymeme; its RST tree always represents the full tweet, not the enthymeme alone.

This yields 756 of 1,334 tweets: 512 easy none cases, that is, nones that all four systems classify correctly, and 244 easy enthymemes. Ease tracks consensus: where the five annotators agree, the models are usually right too (.65 of unanimous nones and .59 of unanimous enthymeme cases are easy); where they split, easy tweets thin out (.26 on even splits). We profile easy nones against the remaining nones and easy enthymemes against the remaining enthymemes within the same vote bands, so the profiles reflect discourse structure rather than vote composition; Table~\ref{tab:easy_profile} reports the enriched and depleted dimensions.

\begin{table}[!tbp]
\centering
\caption{RST dimensions distinguishing the easy tweets, band-matched within each class. Easy and Rest give the mean value of each dimension in the two groups (for binary dimensions, the share of tweets that have it); $\delta$ is Cliff's delta, the strength of the difference: comparing one easy tweet and one rest tweet at random, $\delta$ says how much more often the easy one has the higher value ($0$: no difference; $\pm1$: one group is always higher, whatever the size of the gap, since $\delta$ depends only on the ordering of the values). For binary dimensions $\delta$ equals the difference in shares. Green marks enrichment, red depletion; shading intensity is proportional to $|\delta|$. $p$ is the Mann--Whitney (continuous) or Fisher exact (binary) test; values below $.001$ are collapsed.}
\label{tab:easy_profile}
\footnotesize
\renewcommand{\arraystretch}{1.04}
\begin{tabular*}{\columnwidth}{@{\extracolsep{\fill}} l cc r r @{}}
\toprule
\textbf{Dimension} & \textbf{Easy} & \textbf{Rest} & $\boldsymbol\delta$ & $\boldsymbol{p}$ \\
\midrule
\multicolumn{5}{@{}l}{\textit{Easy nones (512 vs.\ 371)}} \\
\addlinespace[2pt]
Spine ratio & \cellcolor{green!60} .52 & \cellcolor{green!60} .32 & \cellcolor{green!60} $+.48$ & \cellcolor{green!60} $<.001$ \\
Max-relation share & \cellcolor{green!32} .58 & \cellcolor{green!32} .50 & \cellcolor{green!32} $+.16$ & \cellcolor{green!32} $<.001$ \\
Embedded \textsc{joint} & \cellcolor{red!44} .12 & \cellcolor{red!44} .35 & \cellcolor{red!44} $-.22$ & \cellcolor{red!44} $<.001$ \\
\textsc{evaluation} satellite & \cellcolor{red!28} .05 & \cellcolor{red!28} .19 & \cellcolor{red!28} $-.14$ & \cellcolor{red!28} $<.001$ \\
\textsc{attribution} present & \cellcolor{red!24} .21 & \cellcolor{red!24} .33 & \cellcolor{red!24} $-.12$ & \cellcolor{red!24} $<.001$ \\
\textsc{explanation} satellite & \cellcolor{red!17} .04 & \cellcolor{red!17} .12 & \cellcolor{red!17} $-.08$ & \cellcolor{red!17} $.001$ \\
\textsc{cause} satellite$^{a}$ & \cellcolor{red!8} .08 & \cellcolor{red!8} .12 & \cellcolor{red!8} $-.04$ & \cellcolor{red!8} $.048$ \\
\textsc{condition} satellite & \cellcolor{red!7} .03 & \cellcolor{red!7} .07 & \cellcolor{red!7} $-.03$ & \cellcolor{red!7} $.022$ \\
\textsc{contrast} satellite & \cellcolor{red!6} .03 & \cellcolor{red!6} .05 & \cellcolor{red!6} $-.03$ & \cellcolor{red!6} $.045$ \\
\addlinespace[3pt]
\midrule
\multicolumn{5}{@{}l}{\textit{Easy tweets containing an enthymeme (244 vs.\ 207)}} \\
\addlinespace[2pt]
Shallowest-\textsc{arg} span & \cellcolor{green!50} .20 & \cellcolor{green!50} .13 & \cellcolor{green!50} $+.25$ & \cellcolor{green!50} $<.001$ \\
\textsc{arg} share of satellites & \cellcolor{green!48} .27 & \cellcolor{green!48} .18 & \cellcolor{green!48} $+.24$ & \cellcolor{green!48} $<.001$ \\
NN share & \cellcolor{green!32} .27 & \cellcolor{green!32} .22 & \cellcolor{green!32} $+.16$ & \cellcolor{green!32} $.003$ \\
SN share & \cellcolor{green!28} .14 & \cellcolor{green!28} .12 & \cellcolor{green!28} $+.14$ & \cellcolor{green!28} $.006$ \\
Embedded \textsc{joint} & \cellcolor{green!27} .40 & \cellcolor{green!27} .26 & \cellcolor{green!27} $+.14$ & \cellcolor{green!27} $.003$ \\
\textsc{attribution} present & \cellcolor{green!26} .40 & \cellcolor{green!26} .27 & \cellcolor{green!26} $+.13$ & \cellcolor{green!26} $.004$ \\
\textsc{evaluation} satellite & \cellcolor{green!23} .26 & \cellcolor{green!23} .15 & \cellcolor{green!23} $+.12$ & \cellcolor{green!23} $.002$ \\
\textsc{explanation} satellite & \cellcolor{green!18} .18 & \cellcolor{green!18} .09 & \cellcolor{green!18} $+.09$ & \cellcolor{green!18} $.006$ \\
Judgment and reason & \cellcolor{green!17} .15 & \cellcolor{green!17} .06 & \cellcolor{green!17} $+.09$ & \cellcolor{green!17} $.004$ \\
Judgment without reason & \cellcolor{green!15} .25 & \cellcolor{green!15} .17 & \cellcolor{green!15} $+.08$ & \cellcolor{green!15} $.049$ \\
\bottomrule
\addlinespace[2pt]
\multicolumn{5}{@{}p{0.97\columnwidth}@{}}{\footnotesize $^{a}$\,The mildest of the satellite depletions, at .66 of the reference rate.} \\
\end{tabular*}
\end{table}

\subsubsection{Discourse Profile of Easy Nones}
\label{sec:easy_nones}

The 512 easy nones have flat trees. Most of their EDUs are nuclei connected in sequence, with few satellites attached (spine ratio .52 vs.\ .32, $\delta$ $+.48$: the fraction of EDUs that are nuclei on the main path of the tree, without subordinated material). A single relation labels most of the internal nodes, typically \textsc{elaboration} or \textsc{joint} (max-relation share $+.16$: the count of the most frequent relation divided by the number of internal nodes). Argumentative relations are largely absent: satellites of \textsc{cause}, \textsc{explanation}, \textsc{condition}, \textsc{evaluation} or \textsc{contrast} occur at markedly lower rates than in the band-matched reference (half or less for \textsc{explanation}, \textsc{condition} and \textsc{evaluation}), \textsc{attribution} nodes are rare (.21 vs.\ .33), and \textsc{joint} nodes embedded below the root are infrequent (.12 vs.\ .35). 

The easy none is a sequence of nuclei under one repeated relation, with almost no argumentative or attributive satellites, and all four systems classify this configuration correctly.

\subsubsection{Discourse Profile of Easy Tweets Containing an Enthymeme} 
\label{sec:easy_enthymemes}

The 244 easy tweets containing an enthymeme show the opposite configuration: their RST trees contain many argumentative satellites attached close to the root. These satellites of the argumentative relations (\textsc{cause}, \textsc{explanation}, \textsc{condition}, \textsc{evaluation}, \textsc{contrast}, \textsc{comparison}) occur at depth one or two and cover a larger share of the tweet ($+.25$); they also make up a larger fraction of all satellites in its tree ($+.24$). Configurations in which an \textsc{evaluation} or \textsc{contrast} node co-occurs with a \textsc{cause}, \textsc{explanation} or \textsc{condition} node in the same tree are 2.4 times more frequent than in the remaining tweets labeled as potential enthymemes; \textsc{explanation} and \textsc{condition} satellites individually are twice as frequent, \textsc{evaluation} 1.8 times. \textsc{attribution} nodes are 1.5 times more frequent, adjacent EDUs are more often linked by an overt connective, satellite-before-nucleus (SN) order is more frequent, and multinuclear \textsc{joint} nodes embedded below the root are 1.5 times more frequent. 

In these tweets, most components of the argument are expressed as labeled relations in the full-tweet tree, and only part of the argument remains implicit, which is what makes the tweets detectable for all four systems. 

The two profiles differ on the same dimensions in opposite directions: the dimensions less frequent in easy nones are the dimensions more frequent in easy enthymemes, so what all four systems find easy is related to the number and placement of argumentative satellites.

\subsection{Success Analysis by Variation Metric Family}
\label{sec:success_analysis}

\begin{table}[!tbp]
\centering
\caption{Discourse profiles of the four success sets: 22 of the 24 dimensions (the \textsc{condition} and \textsc{comparison} satellite indicators are not shown)
compared against vote-band-matched references (conventions as in
Table~\ref{tab:easy_profile}: set and ref give means, or shares for
binary dimensions; $\delta$ is Cliff's delta, equal to the difference in
shares for binary dimensions; $p$ is the Mann--Whitney or Fisher exact
test, values below $.001$ collapsed). Green marks enrichment, red
depletion; shading intensity is proportional to $|\delta|$ and encodes
effect size, not significance. With set sizes ranging from 15 to 158,
significance tracks sample size as much as effect: small significant
effects in the large sets and large non-significant effects in the small
\textit{where} set should be read accordingly.}
\label{tab:success_profiles}
\scriptsize
\setlength{\tabcolsep}{2pt}
\begin{tabular*}{\textwidth}{@{\extracolsep{\fill}} l cccc cccc cccc cccc @{}}
\toprule
& \multicolumn{4}{c}{\textbf{Where} (15)} & \multicolumn{4}{c}{\textbf{Depth} (158)} & \multicolumn{4}{c}{\textbf{Calibration} (123)} & \multicolumn{4}{c}{\textbf{Consistency} (49)} \\
\cmidrule(lr){2-5} \cmidrule(lr){6-9} \cmidrule(lr){10-13} \cmidrule(lr){14-17}
\textbf{Dimension} & set & ref & $\delta$ & $p$ & set & ref & $\delta$ & $p$ & set & ref & $\delta$ & $p$ & set & ref & $\delta$ & $p$ \\
\midrule
\textsc{attribution} present & \cellcolor{green!57} .53 & \cellcolor{green!57} .25 & \cellcolor{green!57} $+.29$ & \cellcolor{green!57} .031 & \cellcolor{green!5} .28 & \cellcolor{green!5} .25 & \cellcolor{green!5} $+.03$ & \cellcolor{green!5} .48 & \cellcolor{green!5} .26 & \cellcolor{green!5} .24 & \cellcolor{green!5} $+.02$ & \cellcolor{green!5} .73 & \cellcolor{red!5} .24 & \cellcolor{red!5} .25 & \cellcolor{red!5} $-.00$ & \cellcolor{red!5} $>$.99 \\
\textsc{elaboration} share & \cellcolor{red!54} .33 & \cellcolor{red!54} .43 & \cellcolor{red!54} $-.27$ & \cellcolor{red!54} .087 & \cellcolor{green!6} .38 & \cellcolor{green!6} .35 & \cellcolor{green!6} $+.03$ & \cellcolor{green!6} .49 & \cellcolor{green!10} .38 & \cellcolor{green!10} .34 & \cellcolor{green!10} $+.05$ & \cellcolor{green!10} .35 & \cellcolor{green!14} .38 & \cellcolor{green!14} .35 & \cellcolor{green!14} $+.07$ & \cellcolor{green!14} .40 \\
Max-relation share & \cellcolor{red!46} .48 & \cellcolor{red!46} .54 & \cellcolor{red!46} $-.23$ & \cellcolor{red!46} .14 & \cellcolor{green!14} .58 & \cellcolor{green!14} .54 & \cellcolor{green!14} $+.07$ & \cellcolor{green!14} .15 & \cellcolor{green!8} .57 & \cellcolor{green!8} .55 & \cellcolor{green!8} $+.04$ & \cellcolor{green!8} .50 & \cellcolor{green!8} .56 & \cellcolor{green!8} .55 & \cellcolor{green!8} $+.04$ & \cellcolor{green!8} .64 \\
\textsc{attribution} share & \cellcolor{green!46} .12 & \cellcolor{green!46} .07 & \cellcolor{green!46} $+.23$ & \cellcolor{green!46} .071 & \cellcolor{green!6} .11 & \cellcolor{green!6} .09 & \cellcolor{green!6} $+.03$ & \cellcolor{green!6} .38 & \cellcolor{green!5} .11 & \cellcolor{green!5} .09 & \cellcolor{green!5} $+.02$ & \cellcolor{green!5} .60 & \cellcolor{red!5} .08 & \cellcolor{red!5} .10 & \cellcolor{red!5} $-.01$ & \cellcolor{red!5} .88 \\
\textsc{evaluation} satellite & \cellcolor{green!41} .33 & \cellcolor{green!41} .13 & \cellcolor{green!41} $+.20$ & \cellcolor{green!41} .054 & \cellcolor{red!9} .06 & \cellcolor{red!9} .11 & \cellcolor{red!9} $-.05$ & \cellcolor{red!9} .10 & \cellcolor{red!6} .07 & \cellcolor{red!6} .10 & \cellcolor{red!6} $-.03$ & \cellcolor{red!6} .38 & \cellcolor{green!7} .12 & \cellcolor{green!7} .09 & \cellcolor{green!7} $+.04$ & \cellcolor{green!7} .43 \\
Asyndeton & \cellcolor{red!40} .59 & \cellcolor{red!40} .67 & \cellcolor{red!40} $-.20$ & \cellcolor{red!40} .20 & \cellcolor{green!5} .63 & \cellcolor{green!5} .63 & \cellcolor{green!5} $+.02$ & \cellcolor{green!5} .68 & \cellcolor{green!12} .63 & \cellcolor{green!12} .59 & \cellcolor{green!12} $+.06$ & \cellcolor{green!12} .26 & \cellcolor{green!22} .68 & \cellcolor{green!22} .59 & \cellcolor{green!22} $+.11$ & \cellcolor{green!22} .17 \\
Spine ratio & \cellcolor{red!34} .30 & \cellcolor{red!34} .33 & \cellcolor{red!34} $-.17$ & \cellcolor{red!34} .29 & \cellcolor{green!26} .47 & \cellcolor{green!26} .44 & \cellcolor{green!26} $+.13$ & \cellcolor{green!26} .013 & \cellcolor{green!10} .48 & \cellcolor{green!10} .47 & \cellcolor{green!10} $+.05$ & \cellcolor{green!10} .42 & \cellcolor{red!24} .41 & \cellcolor{red!24} .48 & \cellcolor{red!24} $-.12$ & \cellcolor{red!24} .17 \\
SN share & \cellcolor{green!32} .14 & \cellcolor{green!32} .12 & \cellcolor{green!32} $+.16$ & \cellcolor{green!32} .25 & \cellcolor{red!5} .13 & \cellcolor{red!5} .12 & \cellcolor{red!5} $+.00$ & \cellcolor{red!5} .96 & \cellcolor{red!5} .13 & \cellcolor{red!5} .12 & \cellcolor{red!5} $-.01$ & \cellcolor{red!5} .91 & \cellcolor{green!12} .13 & \cellcolor{green!12} .12 & \cellcolor{green!12} $+.06$ & \cellcolor{green!12} .36 \\
\textsc{arg} share of satellites & \cellcolor{green!30} .23 & \cellcolor{green!30} .18 & \cellcolor{green!30} $+.15$ & \cellcolor{green!30} .30 & \cellcolor{red!16} .13 & \cellcolor{red!16} .16 & \cellcolor{red!16} $-.08$ & \cellcolor{red!16} .075 & \cellcolor{red!8} .14 & \cellcolor{red!8} .15 & \cellcolor{red!8} $-.04$ & \cellcolor{red!8} .45 & \cellcolor{green!18} .18 & \cellcolor{green!18} .14 & \cellcolor{green!18} $+.09$ & \cellcolor{green!18} .22 \\
Relation diversity & \cellcolor{green!16} .80 & \cellcolor{green!16} .76 & \cellcolor{green!16} $+.08$ & \cellcolor{green!16} .59 & \cellcolor{green!26} .79 & \cellcolor{green!26} .73 & \cellcolor{green!26} $+.13$ & \cellcolor{green!26} .009 & \cellcolor{green!16} .78 & \cellcolor{green!16} .73 & \cellcolor{green!16} $+.08$ & \cellcolor{green!16} .12 & \cellcolor{green!8} .79 & \cellcolor{green!8} .74 & \cellcolor{green!8} $+.04$ & \cellcolor{green!8} .58 \\
Shallowest-\textsc{arg} span & \cellcolor{green!26} .18 & \cellcolor{green!26} .15 & \cellcolor{green!26} $+.13$ & \cellcolor{green!26} .36 & \cellcolor{red!16} .09 & \cellcolor{red!16} .12 & \cellcolor{red!16} $-.08$ & \cellcolor{red!16} .048 & \cellcolor{red!6} .09 & \cellcolor{red!6} .10 & \cellcolor{red!6} $-.03$ & \cellcolor{red!6} .48 & \cellcolor{green!18} .13 & \cellcolor{green!18} .10 & \cellcolor{green!18} $+.09$ & \cellcolor{green!18} .20 \\
Judgment without reason & \cellcolor{green!22} .27 & \cellcolor{green!22} .16 & \cellcolor{green!22} $+.11$ & \cellcolor{green!22} .28 & \cellcolor{red!11} .10 & \cellcolor{red!11} .15 & \cellcolor{red!11} $-.06$ & \cellcolor{red!11} .075 & \cellcolor{red!10} .08 & \cellcolor{red!10} .13 & \cellcolor{red!10} $-.05$ & \cellcolor{red!10} .16 & \cellcolor{green!5} .12 & \cellcolor{green!5} .12 & \cellcolor{green!5} $+.00$ & \cellcolor{green!5} $>$.99 \\
SN support-first & \cellcolor{green!18} .16 & \cellcolor{green!18} .13 & \cellcolor{green!18} $+.09$ & \cellcolor{green!18} .38 & \cellcolor{red!6} .08 & \cellcolor{red!6} .09 & \cellcolor{red!6} $-.03$ & \cellcolor{red!6} .29 & \cellcolor{red!8} .06 & \cellcolor{red!8} .10 & \cellcolor{red!8} $-.04$ & \cellcolor{red!8} .18 & \cellcolor{red!5} .10 & \cellcolor{red!5} .09 & \cellcolor{red!5} $+.00$ & \cellcolor{red!5} .98 \\
\textsc{joint} at root & \cellcolor{red!17} .00 & \cellcolor{red!17} .08 & \cellcolor{red!17} $-.08$ & \cellcolor{red!17} .60 & \cellcolor{green!5} .09 & \cellcolor{green!5} .08 & \cellcolor{green!5} $+.01$ & \cellcolor{green!5} .63 & \cellcolor{red!8} .05 & \cellcolor{red!8} .09 & \cellcolor{red!8} $-.04$ & \cellcolor{red!8} .19 & \cellcolor{green!5} .08 & \cellcolor{green!5} .08 & \cellcolor{green!5} $+.00$ & \cellcolor{green!5} $>$.99 \\
NN share & \cellcolor{red!6} .21 & \cellcolor{red!6} .21 & \cellcolor{red!6} $-.03$ & \cellcolor{red!6} .85 & \cellcolor{red!10} .20 & \cellcolor{red!10} .21 & \cellcolor{red!10} $-.05$ & \cellcolor{red!10} .33 & \cellcolor{red!10} .18 & \cellcolor{red!10} .20 & \cellcolor{red!10} $-.05$ & \cellcolor{red!10} .40 & \cellcolor{green!16} .23 & \cellcolor{green!16} .19 & \cellcolor{green!16} $+.08$ & \cellcolor{green!16} .28 \\
NS share & \cellcolor{red!5} .65 & \cellcolor{red!5} .66 & \cellcolor{red!5} $-.02$ & \cellcolor{red!5} .93 & \cellcolor{green!5} .58 & \cellcolor{green!5} .57 & \cellcolor{green!5} $+.02$ & \cellcolor{green!5} .70 & \cellcolor{green!10} .59 & \cellcolor{green!10} .56 & \cellcolor{green!10} $+.05$ & \cellcolor{green!10} .36 & \cellcolor{green!16} .62 & \cellcolor{green!16} .56 & \cellcolor{green!16} $+.08$ & \cellcolor{green!16} .37 \\
Embedded \textsc{joint} & \cellcolor{green!5} .33 & \cellcolor{green!5} .32 & \cellcolor{green!5} $+.01$ & \cellcolor{green!5} $>$.99 & \cellcolor{red!12} .16 & \cellcolor{red!12} .22 & \cellcolor{red!12} $-.06$ & \cellcolor{red!12} .12 & \cellcolor{red!8} .15 & \cellcolor{red!8} .19 & \cellcolor{red!8} $-.04$ & \cellcolor{red!8} .37 & \cellcolor{green!13} .24 & \cellcolor{green!13} .18 & \cellcolor{green!13} $+.07$ & \cellcolor{green!13} .25 \\
\textsc{contrast} satellite & \cellcolor{red!13} .00 & \cellcolor{red!13} .06 & \cellcolor{red!13} $-.06$ & \cellcolor{red!13} .60 & \cellcolor{red!5} .03 & \cellcolor{red!5} .04 & \cellcolor{red!5} $-.02$ & \cellcolor{red!5} .49 & \cellcolor{red!5} .02 & \cellcolor{red!5} .04 & \cellcolor{red!5} $-.02$ & \cellcolor{red!5} .39 & \cellcolor{red!7} .00 & \cellcolor{red!7} .04 & \cellcolor{red!7} $-.04$ & \cellcolor{red!7} .39 \\
\textsc{topic--comment} present & \cellcolor{green!8} .13 & \cellcolor{green!8} .09 & \cellcolor{green!8} $+.04$ & \cellcolor{green!8} .64 & \cellcolor{red!5} .04 & \cellcolor{red!5} .06 & \cellcolor{red!5} $-.02$ & \cellcolor{red!5} .43 & \cellcolor{red!5} .04 & \cellcolor{red!5} .04 & \cellcolor{red!5} $-.00$ & \cellcolor{red!5} $>$.99 & \cellcolor{green!13} .10 & \cellcolor{green!13} .04 & \cellcolor{green!13} $+.06$ & \cellcolor{green!13} .050 \\
\textsc{cause} satellite & \cellcolor{green!10} .13 & \cellcolor{green!10} .08 & \cellcolor{green!10} $+.05$ & \cellcolor{green!10} .62 & \cellcolor{red!8} .06 & \cellcolor{red!8} .10 & \cellcolor{red!8} $-.04$ & \cellcolor{red!8} .13 & \cellcolor{green!5} .08 & \cellcolor{green!5} .08 & \cellcolor{green!5} $+.00$ & \cellcolor{green!5} .85 & \cellcolor{green!5} .08 & \cellcolor{green!5} .08 & \cellcolor{green!5} $+.00$ & \cellcolor{green!5} $>$.99 \\
\textsc{explanation} satellite & \cellcolor{red!6} .13 & \cellcolor{red!6} .17 & \cellcolor{red!6} $-.03$ & \cellcolor{red!6} $>$.99 & \cellcolor{red!5} .07 & \cellcolor{red!5} .07 & \cellcolor{red!5} $+.00$ & \cellcolor{red!5} $>$.99 & \cellcolor{red!5} .06 & \cellcolor{red!5} .06 & \cellcolor{red!5} $-.01$ & \cellcolor{red!5} $>$.99 & \cellcolor{green!9} .10 & \cellcolor{green!9} .06 & \cellcolor{green!9} $+.04$ & \cellcolor{green!9} .22 \\
Judgment and reason & \cellcolor{red!5} .07 & \cellcolor{red!5} .07 & \cellcolor{red!5} $-.01$ & \cellcolor{red!5} $>$.99 & \cellcolor{red!8} .03 & \cellcolor{red!8} .07 & \cellcolor{red!8} $-.04$ & \cellcolor{red!8} .071 & \cellcolor{red!5} .04 & \cellcolor{red!5} .06 & \cellcolor{red!5} $-.02$ & \cellcolor{red!5} .40 & \cellcolor{red!9} .02 & \cellcolor{red!9} .06 & \cellcolor{red!9} $-.04$ & \cellcolor{red!9} .35 \\
\bottomrule
\end{tabular*}
\end{table}

Both success poles can be stated in advance. The \textit{where} set contains the genuinely contested tweets, with 2/5 and 3/5 vote splits, on which betabin correctly registers the split; the \textit{depth}, \textit{calibration} and \textit{consistency} sets are almost exclusively unanimous nones (0/5), tweets that all five annotators rejected but on which the text-only model still places near-threshold probabilities. The two poles thus show the same model earning its advantage in opposite directions: restraint where readers divide, commitment where they agree. Table~\ref{tab:success_profiles} gives the full discourse profiles of the four sets.

\subsubsection{Discourse Profile of \textit{Where} Contested Enthymemes}
\label{sec:contested_profiles}

\begin{figure}[!tbp]
\centering
\begin{minipage}[t]{0.485\textwidth}
  \centering
  \includegraphics[width=\linewidth]{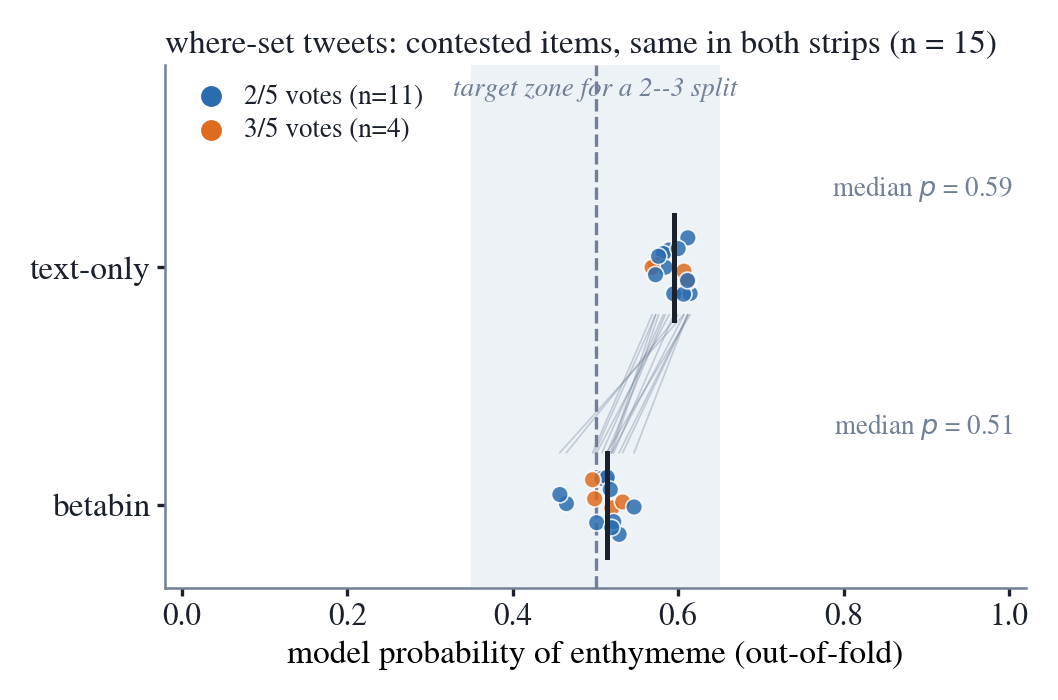}
  \caption{The 15 tweets of the \textit{where} success set (11 with 2/5 votes, 4 with 3/5), same items in both strips; lines connect identical tweets. The shaded band is the target zone for a 2--3 split, whose calibrated probability is near .5. The text-only model clusters at a median of .59; betabin returns the same tweets to a median of .51.}
  \label{fig:where_pole}
\end{minipage}\hfill
\begin{minipage}[t]{0.485\textwidth}
  \centering
  \includegraphics[width=\linewidth]{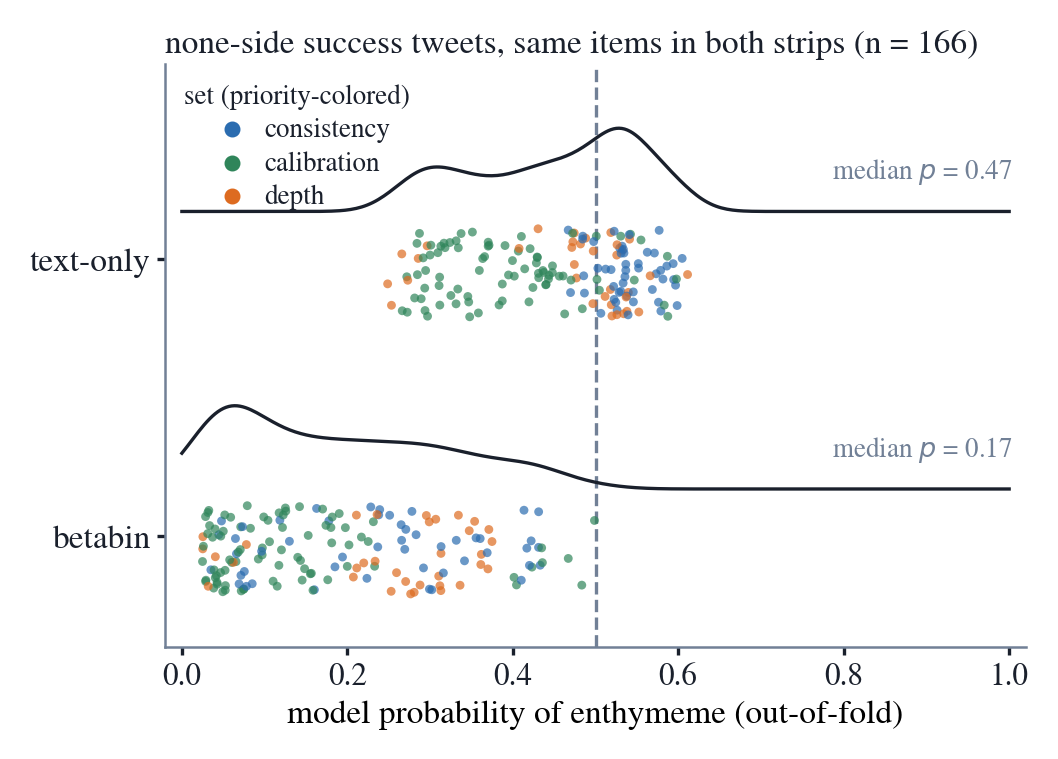}
  \caption{The 166 tweets of the none-side success sets (union of depth, calibration and consistency; colored by set), same items in both strips. Only the horizontal position is informative; points are jittered vertically within a band of fixed height, which gives each strip its rectangular outline, and the curve above each strip is the density of the probabilities. On tweets all five annotators rejected, the text-only model stays at a median of .47, with no tweet below .2; betabin commits, median .17.}
  \label{fig:none_pole_sets}
\end{minipage}
\end{figure}

The \textit{where} set contains 15 tweets, 11 with 2/5 and 4 with 3/5 enthymeme votes, and these alone account for the full ambiguity-Spearman gap (.172 to .209 out-of-fold). Figure~\ref{fig:where_pole} shows both models on these tweets. The text-only model sits confidently on one side, a tight cluster at .55 to .62; since 11 of the 15 are majority-none items, it is on the F1 wrong side of the split for most of them. Betabin moves every one of these tweets toward .5. Both models remain inside the loose target zone, so the gain is not absolute calibration but ranking: a probability of .60 on a contested tweet ranks it as less uncertain than the model's own hedged unanimous nones, breaking the uncertainty ordering, whereas betabin's .51 here, against its committed nones near .07, places the contested tweets at the top of its uncertainty ranking, where the annotators place them.

The discourse profile of these 15 tweets, compared with the 109 other contested items, points to a single shared configuration: \textsc{attribution} nodes are twice as frequent (.53 vs.\ .25, $p=.031$), \textsc{evaluation} satellites 2.6 times more frequent (.33 vs.\ .13, $p=.054$), and \textsc{elaboration} share is lower ($\delta$ $-.27$). A structural backbone emerges: an \textsc{attribution} node paired with an \textsc{evaluation} satellite, that is, reported speech from an exterior source together with a judgment of it. An example tree from this set is shown in Figure~\ref{fig:rst_1965} of Appendix~\ref{sec:appendix_rst_trees}.

This configuration generates both types of disagreement in enthymeme annotation identified in \citet{pastor-vanarkel-2026-disagreement}. Where the \textsc{attribution} nucleus points to material outside the tweet, orphaned references and bridging inferences make resolution possible but costly, and annotators are more prone to inconsistent annotation. Then, in this set, the evaluation is often ironic or framed as a rhetorical question: no retrieval settles whether an implicit premise is advanced, and the divergence reflects subjective perception of the speech act, corresponding to genuine disagreement. Most of the fifteen cases admit both readings, while sharing the same discourse structure backbone which is why they are maximally contested. 

The models hence divide accordingly: the text-only model reads the \textsc{attribution}--\allowbreak\textsc{evaluation} surface as enthymematic and commits (median .59); betabin has learned that this configuration produces split votes, and remains at .51.\footnote{With $n=15$, we read this profile as convergent evidence together with Section~\ref{sec:probing}, not as standalone significance.}

\subsubsection{Discourse Profile of \textit{Depth, Calibration, Consistency} Unanimous Nones}
\label{sec:consistency_profiles}

The remaining sets of tweets from the three other metrics are entirely different. The depth set contains 158 tweets, of which 154 are unanimous nones; the calibration set (123) and the consistency set (49) are unanimous nones entirely, and the three overlap heavily, with a 40-tweet core common to all of them. No 3/5 4/5 5/5 enthymeme appears in any set. 

Figure~\ref{fig:none_pole_sets} shows the union: on tweets that received zero enthymeme votes, the text-only model places a median probability of .47, with 77\% of the set at the threshold and none committed, while betabin commits on 57\% of them, median .17. The pattern holds for the full population of 608 unanimous nones. 

The discourse profiles at this pole are real but less pronounced than for the easy nones of Table~\ref{tab:easy_profile}. Against the band-matched reference, the depth set shows a weaker version of the easy-none configuration (spine ratio $\delta$ $+.13$, $p=.013$; relation diversity $\delta$ $+.13$, $p=.009$). A direct comparison with the easy unanimous nones indicates however the tweets on which the text-only model stays at the threshold contain residual argumentative structure, with more EDUs (3.7 vs.\ 3.0, $p<.001$), larger argumentative satellites ($\delta$ $+.11$, $p=.008$), more \textsc{attribution} nodes (.28 vs.\ .18, $p=.016$) and a lower spine ratio ($\delta$ $-.14$, $p=.010$). It seems this residual structure is sufficient to hold the text-only model at the threshold, but not sufficient for the annotators to identify an enthymeme. Betabin, whose training objective penalizes near-threshold predictions on unanimous items, commits despite it. The calibration set shows no enriched dimension at all, indicating that betabin's calibration gain does not depend on discourse structure.

The consistency set shows one enriched dimension, \textsc{topic--comment} framing (.10 vs.\ .04, $p=.050$), alongside depleted \textsc{joint} presence and a higher single-relation share. The effects are marginal, but the frame is telling: its subtypes include rhetorical questions and comments, which \citet{pastor-vanarkel-2026-disagreement} identify as a source of spurious disagreement, since converting a rhetorical question into a proposition is a reconstruction step the guidelines do not prescribe, so that whether an annotator performs it varies from one annotator to another and from one tweet to the next. These are the nones most likely to draw a stray vote. Figure~\ref{fig:none_pole_sets} shows the corresponding behaviors: the text-only model holds these tweets just below the threshold, registering the ambiguity without resolving it, while betabin moves them to the none side with visibly different degrees of commitment. Because betabin models the individual votes rather than the majority label, it learns how likely a reader is to reconstruct, or fail to reconstruct, the argument behind such a frame, and so measures the possibility of an inconsistent vote itself. Only the per-vote consistency metric can observe this, since a majority-label metric registers the stray vote as noise and a single threshold value cannot represent a doubt that is not uniform across tweets. Figure~\ref{fig:rst_481} of Appendix~\ref{sec:appendix_rst_trees} shows one such tree.

Taken together, the two poles show the same model obtaining its advantage through opposite behaviors: it commits where the annotators are unanimous and remains near the threshold where they divide. The discourse profiles indicate that each behavior corresponds to an identifiable structure: the absence of argumentative satellites at the `none' pole, reported speech with evaluative satellites in the contested middle. A model with a single compressed probability scale exhibits neither behavior, and a metric computed against the majority label registers neither. 

The asymmetry also introduces the error analysis: betabin never commits at the enthymeme pole, and we argue that this is systematic since the certain absence of an enthymeme is identifiable from discourse structure while its certain presence is not.

\subsection{Error Analysis by Variation Metric Family}
\label{sec:error_cases}

We organize this error analysis by the metric families of Section~\ref{sec:metrics}, since each family isolates a different model behavior and, with it, a different set of tweets. We begin with the \textit{where} family, which covers both the contested band and the consensus poles. We then treat the three remaining families together, \textit{depth}, \textit{calibration} and \textit{consistency}, because their success sets are not independent: the tweets they single out nest into one another, from the broad depth set down to the small consistency set, so the three are best read as successive refinements of one behavior at the consensus poles rather than as three separate phenomena.

\begin{table}[!tbp]
\centering
\caption{Discourse profile of the \textit{where}-failure set --- the 120
tweets whose contestedness every system misjudges (uncertainty rank
displaced from the vote-entropy rank by more than 30 percentile points in all
four transformers) --- split by vote band and compared against the
remaining tweets of the same bands (same conventions as in previous Tables).}
\label{tab:where_failure_profile}
\scriptsize
\setlength{\tabcolsep}{2pt}
\begin{tabular*}{\textwidth}{@{\extracolsep{\fill}} l cccc cccc cccc @{}}
\toprule
& \multicolumn{4}{c}{\textbf{Consensus none}} & \multicolumn{4}{c}{\textbf{Contested}} & \multicolumn{4}{c}{\textbf{Consensus enth.}} \\
& \multicolumn{4}{c}{68 (32$\times$0/5, 36$\times$1/5)} & \multicolumn{4}{c}{18 (11$\times$2/5, 7$\times$3/5)} & \multicolumn{4}{c}{34 (13$\times$4/5, 21$\times$5/5)} \\
\cmidrule(lr){2-5} \cmidrule(lr){6-9} \cmidrule(lr){10-13}
\textbf{Dimension} & set & ref & $\delta$ & $p$ & set & ref & $\delta$ & $p$ & set & ref & $\delta$ & $p$ \\
\midrule
Spine ratio & \cellcolor{red!5} .42 & \cellcolor{red!5} .45 & \cellcolor{red!5} $-.03$ & \cellcolor{red!5} .72 & \cellcolor{green!60} .42 & \cellcolor{green!60} .31 & \cellcolor{green!60} $+.34$ & \cellcolor{green!60} .022 & \cellcolor{green!42} .40 & \cellcolor{green!42} .33 & \cellcolor{green!42} $+.21$ & \cellcolor{green!42} .042 \\
\textsc{attribution} present & \cellcolor{green!5} .27 & \cellcolor{green!5} .26 & \cellcolor{green!5} $+.01$ & \cellcolor{green!5} .89 & \cellcolor{red!53} .06 & \cellcolor{red!53} .32 & \cellcolor{red!53} $-.26$ & \cellcolor{red!53} .022 & \cellcolor{red!38} .18 & \cellcolor{red!38} .37 & \cellcolor{red!38} $-.19$ & \cellcolor{red!38} .037 \\
Max-relation share & \cellcolor{green!11} .57 & \cellcolor{green!11} .54 & \cellcolor{green!11} $+.06$ & \cellcolor{green!11} .44 & \cellcolor{green!12} .56 & \cellcolor{green!12} .53 & \cellcolor{green!12} $+.06$ & \cellcolor{green!12} .67 & \cellcolor{green!50} .59 & \cellcolor{green!50} .48 & \cellcolor{green!50} $+.25$ & \cellcolor{green!50} .014 \\
\textsc{attribution} share & \cellcolor{green!5} .12 & \cellcolor{green!5} .09 & \cellcolor{green!5} $+.02$ & \cellcolor{green!5} .71 & \cellcolor{red!50} .03 & \cellcolor{red!50} .09 & \cellcolor{red!50} $-.25$ & \cellcolor{red!50} .033 & \cellcolor{red!34} .06 & \cellcolor{red!34} .10 & \cellcolor{red!34} $-.17$ & \cellcolor{red!34} .053 \\
Relation diversity & \cellcolor{green!5} .77 & \cellcolor{green!5} .74 & \cellcolor{green!5} $+.02$ & \cellcolor{green!5} .82 & \cellcolor{green!41} .81 & \cellcolor{green!41} .76 & \cellcolor{green!41} $+.20$ & \cellcolor{green!41} .15 & \cellcolor{green!27} .82 & \cellcolor{green!27} .77 & \cellcolor{green!27} $+.13$ & \cellcolor{green!27} .19 \\
\textsc{elaboration} share & \cellcolor{green!5} .36 & \cellcolor{green!5} .35 & \cellcolor{green!5} $+.01$ & \cellcolor{green!5} .84 & \cellcolor{green!17} .47 & \cellcolor{green!17} .41 & \cellcolor{green!17} $+.08$ & \cellcolor{green!17} .57 & \cellcolor{green!37} .45 & \cellcolor{green!37} .33 & \cellcolor{green!37} $+.19$ & \cellcolor{green!37} .069 \\
Embedded \textsc{joint} & \cellcolor{red!6} .18 & \cellcolor{red!6} .21 & \cellcolor{red!6} $-.03$ & \cellcolor{red!6} .64 & \cellcolor{red!36} .17 & \cellcolor{red!36} .35 & \cellcolor{red!36} $-.18$ & \cellcolor{red!36} .17 & \cellcolor{green!5} .35 & \cellcolor{green!5} .34 & \cellcolor{green!5} $+.01$ & \cellcolor{green!5} $>$.99 \\
Asyndeton & \cellcolor{green!32} .72 & \cellcolor{green!32} .63 & \cellcolor{green!32} $+.16$ & \cellcolor{green!32} .024 & \cellcolor{red!5} .63 & \cellcolor{red!5} .67 & \cellcolor{red!5} $-.02$ & \cellcolor{red!5} .92 & \cellcolor{green!34} .74 & \cellcolor{green!34} .67 & \cellcolor{green!34} $+.17$ & \cellcolor{green!34} .097 \\
SN share & \cellcolor{red!9} .14 & \cellcolor{red!9} .13 & \cellcolor{red!9} $-.04$ & \cellcolor{red!9} .47 & \cellcolor{red!27} .09 & \cellcolor{red!27} .12 & \cellcolor{red!27} $-.14$ & \cellcolor{red!27} .29 & \cellcolor{red!10} .13 & \cellcolor{red!10} .13 & \cellcolor{red!10} $-.05$ & \cellcolor{red!10} .61 \\
\textsc{explanation} satellite & \cellcolor{green!5} .09 & \cellcolor{green!5} .07 & \cellcolor{green!5} $+.02$ & \cellcolor{green!5} .47 & \cellcolor{red!25} .06 & \cellcolor{red!25} .18 & \cellcolor{red!25} $-.12$ & \cellcolor{red!25} .30 & \cellcolor{red!17} .06 & \cellcolor{red!17} .15 & \cellcolor{red!17} $-.09$ & \cellcolor{red!17} .20 \\
NS share & \cellcolor{green!13} .60 & \cellcolor{green!13} .57 & \cellcolor{green!13} $+.07$ & \cellcolor{green!13} .36 & \cellcolor{green!20} .68 & \cellcolor{green!20} .66 & \cellcolor{green!20} $+.10$ & \cellcolor{green!20} .50 & \cellcolor{green!8} .62 & \cellcolor{green!8} .61 & \cellcolor{green!8} $+.04$ & \cellcolor{green!8} .69 \\
Judgment and reason & \cellcolor{red!5} .04 & \cellcolor{red!5} .07 & \cellcolor{red!5} $-.02$ & \cellcolor{red!5} .61 & \cellcolor{red!17} .00 & \cellcolor{red!17} .09 & \cellcolor{red!17} $-.08$ & \cellcolor{red!17} .35 & \cellcolor{red!18} .03 & \cellcolor{red!18} .12 & \cellcolor{red!18} $-.09$ & \cellcolor{red!18} .15 \\
\textsc{evaluation} satellite & \cellcolor{red!6} .07 & \cellcolor{red!6} .10 & \cellcolor{red!6} $-.03$ & \cellcolor{red!6} .53 & \cellcolor{red!10} .11 & \cellcolor{red!10} .16 & \cellcolor{red!10} $-.05$ & \cellcolor{red!10} .74 & \cellcolor{red!16} .15 & \cellcolor{red!16} .23 & \cellcolor{red!16} $-.08$ & \cellcolor{red!16} .39 \\
NN share & \cellcolor{red!5} .20 & \cellcolor{red!5} .21 & \cellcolor{red!5} $-.02$ & \cellcolor{red!5} .72 & \cellcolor{red!15} .18 & \cellcolor{red!15} .22 & \cellcolor{red!15} $-.08$ & \cellcolor{red!15} .59 & \cellcolor{red!7} .26 & \cellcolor{red!7} .25 & \cellcolor{red!7} $-.04$ & \cellcolor{red!7} .73 \\
\textsc{cause} satellite & \cellcolor{green!5} .10 & \cellcolor{green!5} .10 & \cellcolor{green!5} $+.01$ & \cellcolor{green!5} .83 & \cellcolor{green!5} .11 & \cellcolor{green!5} .09 & \cellcolor{green!5} $+.03$ & \cellcolor{green!5} .66 & \cellcolor{green!15} .21 & \cellcolor{green!15} .13 & \cellcolor{green!15} $+.07$ & \cellcolor{green!15} .29 \\
\textsc{arg} share of satellites & \cellcolor{red!5} .15 & \cellcolor{red!5} .16 & \cellcolor{red!5} $-.01$ & \cellcolor{red!5} .82 & \cellcolor{red!5} .22 & \cellcolor{red!5} .18 & \cellcolor{red!5} $-.01$ & \cellcolor{red!5} .96 & \cellcolor{red!13} .23 & \cellcolor{red!13} .23 & \cellcolor{red!13} $-.07$ & \cellcolor{red!13} .50 \\
\textsc{contrast} satellite & \cellcolor{red!5} .03 & \cellcolor{red!5} .04 & \cellcolor{red!5} $-.01$ & \cellcolor{red!5} $>$.99 & \cellcolor{green!13} .11 & \cellcolor{green!13} .05 & \cellcolor{green!13} $+.06$ & \cellcolor{green!13} .27 & \cellcolor{red!5} .03 & \cellcolor{red!5} .04 & \cellcolor{red!5} $-.01$ & \cellcolor{red!5} $>$.99 \\
Judgment without reason & \cellcolor{red!8} .10 & \cellcolor{red!8} .14 & \cellcolor{red!8} $-.04$ & \cellcolor{red!8} .47 & \cellcolor{green!12} .22 & \cellcolor{green!12} .16 & \cellcolor{green!12} $+.06$ & \cellcolor{green!12} .51 & \cellcolor{red!9} .18 & \cellcolor{red!9} .22 & \cellcolor{red!9} $-.05$ & \cellcolor{red!9} .67 \\
SN support-first & \cellcolor{red!11} .04 & \cellcolor{red!11} .09 & \cellcolor{red!11} $-.06$ & \cellcolor{red!11} .16 & \cellcolor{red!12} .11 & \cellcolor{red!12} .14 & \cellcolor{red!12} $-.06$ & \cellcolor{red!12} .55 & \cellcolor{green!7} .14 & \cellcolor{green!7} .13 & \cellcolor{green!7} $+.04$ & \cellcolor{green!7} .60 \\
\textsc{topic--comment} present & \cellcolor{red!5} .03 & \cellcolor{red!5} .05 & \cellcolor{red!5} $-.02$ & \cellcolor{red!5} .57 & \cellcolor{red!10} .06 & \cellcolor{red!10} .10 & \cellcolor{red!10} $-.05$ & \cellcolor{red!10} $>$.99 & \cellcolor{red!11} .06 & \cellcolor{red!11} .11 & \cellcolor{red!11} $-.06$ & \cellcolor{red!11} .56 \\
Shallowest-\textsc{arg} span & \cellcolor{green!5} .13 & \cellcolor{green!5} .11 & \cellcolor{green!5} $+.00$ & \cellcolor{green!5} .95 & \cellcolor{red!6} .14 & \cellcolor{red!6} .15 & \cellcolor{red!6} $-.03$ & \cellcolor{red!6} .83 & \cellcolor{red!11} .18 & \cellcolor{red!11} .17 & \cellcolor{red!11} $-.05$ & \cellcolor{red!11} .58 \\
\bottomrule
\end{tabular*}
\end{table}

\subsubsection{Discourse Profile of \textit{Where} Failures}
\label{sec:where_failures}
\label{sec:evidence_elaboration}

The \textit{where} metric concerns the ranking of tweets by uncertainty rather than the placement of a probability. It only scores an ordering, asking whether the tweets the model is least certain about, those with probabilities near .5, are the tweets with contested vote counts rather than unanimous ones. A model can thus place its probabilities close to the vote fractions and still order the tweets wrongly, and conversely order them correctly while misplacing many probabilities. Its failure set contains the 120 tweets whose position in this ordering all four systems misjudge, profiled by vote band in Table~\ref{tab:where_failure_profile} (68 consensus nones, 18 contested, 34 consensus enthymemes).

The misranked consensus nones show a single enrichment in terms of discourse dimensions. Relative to the other tweets in the same vote bands, they have the highest share of implicitly marked relations (asyndeton .72 vs.\ .63, $p=.024$), while on every other dimension they do not differ, with $|\delta| \leq .07$ throughout. Implicit marking is itself an ambiguous cue for enthymemes, which may be what moves these tweets toward ambiguity. The same enrichment appears in the full consensus settled enthymemes below (.74 vs.\ .67), where the unmarked junctions conceal \textsc{evidence} relations and their unstated warrants; on these nones, the same surface configuration conceals nothing. An unmarked junction is compatible with both readings, and a model that has learned to treat implicit marking as a signal of possible implicitness will misrank precisely the consensus nones that happen to be least marked.

The misranked contested tweets carry a high spine ratio ($+.34$, $p=.022$) and almost no reported speech, with an \textsc{attribution} node in 1 of the 18 (.06 vs.\ .32, $p=.022$); all four systems place them among their most confident nones. Furthermore, the comparison with the \textit{where}-success set, drawn from the same 124 contested items, yields a sharp distinction, since the correctly ranked contested tweets carry \textsc{attribution} at .53 and the misranked ones at .06. Reported speech thus functions as the models' cue for contestedness, and a contested tweet without an exterior source tends to be classified towards the plain nones. 

The full consensus settled enthymemes on which no system commits show the same configuration in a stronger form, with one relation dominating the tree (max-relation share $+.25$, $p=.014$; \textsc{elaboration} share $+.19$, $p=.069$), a high spine ratio ($+.21$, $p=.042$), depleted reported speech ($-.19$, $p=.037$), and a high share of implicitly marked relations ($+.17$, $p=.097$). Four or five annotators recognized these tweets as arguments; every model treats them as contested. This set and the misranked contested set share the same profile, a tree with a high spine ratio and little reported speech, while the annotators behave differently on them, splitting on the former and affirming the latter. Since the models observe only this shared surface, genuinely contested tweets and unanimously affirmed enthymemes fall into a single probability region and these consensus enthymemes are never fully commited.

We suggest that these full consensus enthymeme failures are structural rather than reducible by more training data, and that the implicit marking of coherence relations is what makes them so. In these misranked full consensus enthymemes, the relations between EDUs are not explicitly signaled, and the parser annotates the unmarked junctions as \textsc{elaboration} or \textsc{joint}. This is a parser mistake, since the operative relations by the criteria of \citet{mann1988rhetorical} are semantic ones such as \textsc{evidence}, \textsc{cause} or \textsc{explanation}, and where these relations are not explicitly signaled, an unstated warrant supplied by the reader is very often in place. 

\begin{figure}[!tbp]
    \centering
    \includegraphics[width=0.55\textwidth]{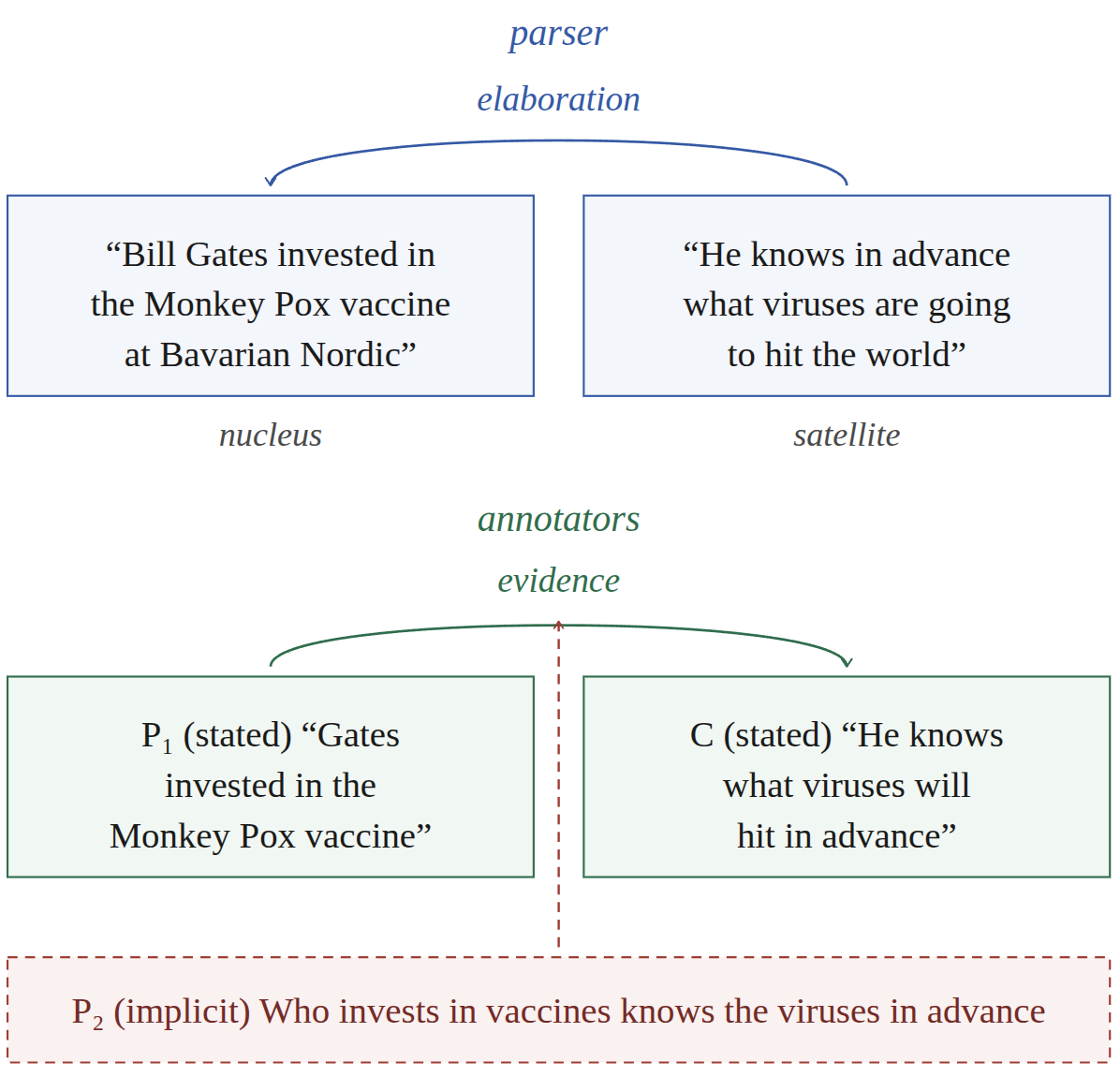}
    \caption{The two construals of the unmarked junction on Tweet \#196 (2/5 votes). The parser reads \textsc{elaboration}, a continuation; the annotators who hold the implicit warrant $P_2$ read \textsc{evidence}. The structure fixes the site of the argument; whether it is read depends on the reader supplying $P_2$.}
    \label{fig:evidence_elaboration}
\end{figure}

Even though only some of our models receive the parsed RST features as inputs, all of them read the same unmarked surface and make the same mistake, treating these junctions as continuation\footnote{Continuation is to be understood here as an additive relation in the taxonomy of \citet{sanders1992toward}, mere conjunction of the two segments, as opposed to the pragmatic causal relations such as \textsc{evidence}, in which one segment bears on the acceptance of the other.} where some annotators identified an argument. 

What separates the two readings is not in the text. The junction carries indeed an evidence relation no matter what, but reading it requires supplying the warrant that connects the two spans. For a reader who holds that warrant the inference fires without being noticed, since the connection is sedimented; a reader who lacks it registers a gap and can question what the junction claims. Whether a reader holds the warrant depends on their background, on what circulates as accepted in their community. The parser (or enthymeme detection system) holds nothing, and mistakes the unmarked junction for mere continuation. Figure~\ref{fig:evidence_elaboration} shows the two readings on Tweet \#196, where the warrant to supply is conspiratorial : ''Imagine that Bill Gates invested in the Monkey Pox vaccine at Bavarian Nordic --- Wow --- He knows in advance what viruses are going to hit the world --- Imagine that,'' received two enthymeme votes out of five.

\begin{table}[!tbp]
\centering
\caption{The two consensus-pole failure profiles, extracted from the
nested \textit{depth}, \textit{calibration} and \textit{consistency}
failure sets. Shown is the no-commitment criterion applied to the
consensus bands: tweets on which no system places a probability within
$.35$ of the majority pole, profiled against the committed tweets of the
same bands (conventions as in Table~\ref{tab:easy_profile}; $\delta$ is
Cliff's delta; $p$ is Mann--Whitney / Fisher exact; green marks
enrichment, red depletion, shading proportional to $|\delta|$;
uncorrected screening).}
\label{tab:failure_two_profiles}
\scriptsize
\setlength{\tabcolsep}{7pt}
\begin{tabular*}{\textwidth}{@{\extracolsep{\fill}} l cccc cccc @{}}
\toprule
& \multicolumn{4}{c}{\textbf{Uncommitted consensus nones}} & \multicolumn{4}{c}{\textbf{Uncommitted consensus enth.}} \\
& \multicolumn{4}{c}{299 (185$\times$0/5, 114$\times$1/5) vs.\ 527} & \multicolumn{4}{c}{313 (131$\times$4/5, 182$\times$5/5) vs.\ 71} \\
\cmidrule(lr){2-5} \cmidrule(lr){6-9}
\textbf{Dimension} & set & ref & $\delta$ & $p$ & set & ref & $\delta$ & $p$ \\
\midrule
Spine ratio & \cellcolor{red!60} .32 & \cellcolor{red!60} .51 & \cellcolor{red!60} $-.45$ & \cellcolor{red!60} $<$.001 & \cellcolor{green!13} .34 & \cellcolor{green!13} .31 & \cellcolor{green!13} $+.06$ & \cellcolor{green!13} .41 \\
Shallowest-\textsc{arg} span & \cellcolor{green!58} .17 & \cellcolor{green!58} .08 & \cellcolor{green!58} $+.29$ & \cellcolor{green!58} $<$.001 & \cellcolor{red!28} .17 & \cellcolor{red!28} .18 & \cellcolor{red!28} $-.14$ & \cellcolor{red!28} .053 \\
\textsc{arg} share of satellites & \cellcolor{green!57} .23 & \cellcolor{green!57} .12 & \cellcolor{green!57} $+.29$ & \cellcolor{green!57} $<$.001 & \cellcolor{red!46} .21 & \cellcolor{red!46} .32 & \cellcolor{red!46} $-.23$ & \cellcolor{red!46} .001 \\
NN share & \cellcolor{green!44} .25 & \cellcolor{green!44} .18 & \cellcolor{green!44} $+.22$ & \cellcolor{green!44} $<$.001 & \cellcolor{red!28} .25 & \cellcolor{red!28} .29 & \cellcolor{red!28} $-.14$ & \cellcolor{red!28} .060 \\
Max-relation share & \cellcolor{red!34} .50 & \cellcolor{red!34} .57 & \cellcolor{red!34} $-.17$ & \cellcolor{red!34} $<$.001 & \cellcolor{green!42} .51 & \cellcolor{green!42} .42 & \cellcolor{green!42} $+.21$ & \cellcolor{green!42} .005 \\
Asyndeton & \cellcolor{green!6} .68 & \cellcolor{green!6} .61 & \cellcolor{green!6} $+.03$ & \cellcolor{green!6} .46 & \cellcolor{green!37} .69 & \cellcolor{green!37} .63 & \cellcolor{green!37} $+.19$ & \cellcolor{green!37} .013 \\
Embedded \textsc{joint} & \cellcolor{green!36} .32 & \cellcolor{green!36} .14 & \cellcolor{green!36} $+.18$ & \cellcolor{green!36} $<$.001 & \cellcolor{red!16} .33 & \cellcolor{red!16} .41 & \cellcolor{red!16} $-.08$ & \cellcolor{red!16} .21 \\
\textsc{evaluation} satellite & \cellcolor{green!25} .18 & \cellcolor{green!25} .06 & \cellcolor{green!25} $+.13$ & \cellcolor{green!25} $<$.001 & \cellcolor{red!29} .19 & \cellcolor{red!29} .34 & \cellcolor{red!29} $-.14$ & \cellcolor{red!29} .011 \\
Judgment without reason & \cellcolor{green!27} .23 & \cellcolor{green!27} .09 & \cellcolor{green!27} $+.14$ & \cellcolor{green!27} $<$.001 & \cellcolor{red!15} .20 & \cellcolor{red!15} .28 & \cellcolor{red!15} $-.08$ & \cellcolor{red!15} .16 \\
\textsc{elaboration} share & \cellcolor{green!7} .35 & \cellcolor{green!7} .36 & \cellcolor{green!7} $+.04$ & \cellcolor{green!7} .36 & \cellcolor{green!23} .36 & \cellcolor{green!23} .29 & \cellcolor{green!23} $+.12$ & \cellcolor{green!23} .12 \\
Relation diversity & \cellcolor{red!17} .76 & \cellcolor{red!17} .74 & \cellcolor{red!17} $-.09$ & \cellcolor{red!17} .028 & \cellcolor{red!19} .77 & \cellcolor{red!19} .81 & \cellcolor{red!19} $-.10$ & \cellcolor{red!19} .19 \\
Judgment and reason & \cellcolor{green!18} .12 & \cellcolor{green!18} .03 & \cellcolor{green!18} $+.09$ & \cellcolor{green!18} $<$.001 & \cellcolor{red!14} .10 & \cellcolor{red!14} .17 & \cellcolor{red!14} $-.07$ & \cellcolor{red!14} .098 \\
\textsc{explanation} satellite & \cellcolor{green!18} .13 & \cellcolor{green!18} .04 & \cellcolor{green!18} $+.09$ & \cellcolor{green!18} $<$.001 & \cellcolor{red!18} .12 & \cellcolor{red!18} .21 & \cellcolor{red!18} $-.09$ & \cellcolor{red!18} .057 \\
NS share & \cellcolor{green!18} .62 & \cellcolor{green!18} .54 & \cellcolor{green!18} $+.09$ & \cellcolor{green!18} .032 & \cellcolor{green!17} .62 & \cellcolor{green!17} .58 & \cellcolor{green!17} $+.09$ & \cellcolor{green!17} .25 \\
\textsc{topic--comment} present & \cellcolor{green!17} .11 & \cellcolor{green!17} .02 & \cellcolor{green!17} $+.09$ & \cellcolor{green!17} $<$.001 & \cellcolor{red!5} .11 & \cellcolor{red!5} .13 & \cellcolor{red!5} $-.02$ & \cellcolor{red!5} .67 \\
SN share & \cellcolor{green!17} .12 & \cellcolor{green!17} .13 & \cellcolor{green!17} $+.09$ & \cellcolor{green!17} .015 & \cellcolor{red!5} .13 & \cellcolor{red!5} .13 & \cellcolor{red!5} $-.02$ & \cellcolor{red!5} .72 \\
SN support-first & \cellcolor{green!17} .13 & \cellcolor{green!17} .07 & \cellcolor{green!17} $+.08$ & \cellcolor{green!17} $<$.001 & \cellcolor{red!11} .12 & \cellcolor{red!11} .17 & \cellcolor{red!11} $-.06$ & \cellcolor{red!11} .26 \\
\textsc{attribution} present & \cellcolor{green!16} .31 & \cellcolor{green!16} .23 & \cellcolor{green!16} $+.08$ & \cellcolor{green!16} .013 & \cellcolor{red!11} .34 & \cellcolor{red!11} .39 & \cellcolor{red!11} $-.06$ & \cellcolor{red!11} .41 \\
\textsc{cause} satellite & \cellcolor{green!11} .13 & \cellcolor{green!11} .08 & \cellcolor{green!11} $+.05$ & \cellcolor{green!11} .013 & \cellcolor{green!5} .14 & \cellcolor{green!5} .13 & \cellcolor{green!5} $+.01$ & \cellcolor{green!5} .85 \\
\textsc{condition} satellite & \cellcolor{green!10} .08 & \cellcolor{green!10} .03 & \cellcolor{green!10} $+.05$ & \cellcolor{green!10} .001 & \cellcolor{green!5} .07 & \cellcolor{green!5} .07 & \cellcolor{green!5} $+.00$ & \cellcolor{green!5} $>$.99 \\
\textsc{attribution} share & \cellcolor{green!9} .08 & \cellcolor{green!9} .11 & \cellcolor{green!9} $+.04$ & \cellcolor{green!9} .18 & \cellcolor{red!5} .09 & \cellcolor{red!5} .08 & \cellcolor{red!5} $-.01$ & \cellcolor{red!5} .86 \\
\textsc{contrast} satellite & \cellcolor{green!7} .06 & \cellcolor{green!7} .02 & \cellcolor{green!7} $+.04$ & \cellcolor{green!7} .013 & \cellcolor{red!5} .04 & \cellcolor{red!5} .04 & \cellcolor{red!5} $-.01$ & \cellcolor{red!5} .73 \\
\textsc{joint} at root & \cellcolor{green!5} .09 & \cellcolor{green!5} .08 & \cellcolor{green!5} $+.01$ & \cellcolor{green!5} .59 & \cellcolor{green!7} .06 & \cellcolor{green!7} .03 & \cellcolor{green!7} $+.03$ & \cellcolor{green!7} .39 \\
\textsc{comparison} satellite & \cellcolor{green!5} .01 & \cellcolor{green!5} .00 & \cellcolor{green!5} $+.01$ & \cellcolor{green!5} .36 & \cellcolor{green!5} .00 & \cellcolor{green!5} .00 & \cellcolor{green!5} $+.00$ & \cellcolor{green!5} $>$.99 \\
\bottomrule
\end{tabular*}
\end{table}

\subsubsection{Discourse Profile of \textit{Depth, Calibration, Consistency} Failures}
\label{sec:pole_failures}

The \textit{depth}, \textit{calibration} and \textit{consistency} failures are not three populations but nested sets at both consensus poles ($\text{consistency} \subset \text{calibration} \subset \text{depth}$). We therefore give their discourse profile all at once, through the commitment criterion of Table~\ref{tab:failure_two_profiles}, comparing consensus tweets on which no system places a probability within .35 of the majority pole against the committed tweets of the same bands. The poles are strongly asymmetric, with 299 of 826 consensus nones uncommitted (36\%) against 313 of 384 consensus enthymemes (81\%).

First, the uncommitted consensus nones are nones with argument structure. Their trees branch rather than chain (spine ratio .32 vs.\ .51, $\delta=-.45$, $p<.001$), and every argumentative dimension is enriched, with the share of argumentative satellites at double the reference (.23 vs.\ .12, $p<.001$) and close to the root (shallowest-\textsc{arg} span $+.29$), \textsc{evaluation} (.18 vs.\ .06) and \textsc{explanation} (.13 vs.\ .04) satellites, judgments with and without a stated reason, \textsc{topic--comment} structures (.11 vs.\ .02) and embedded \textsc{joint} nodes (.32 vs.\ .14), all at $p<.001$. This is the configuration of the easy enthymemes of Table~\ref{tab:easy_profile}. These tweets state a judgment and support it; what they do not do is omit a premise. The models place near-threshold probabilities on precisely the nones that carry the surface structure of argument, and implicit marking plays no role on this side (asyndeton $+.03$, n.s.).

The uncommitted consensus enthymemes show the configuration of the misranked settled enthymemes of the \textit{where} analysis, now at scale (313 tweets against 34). Argumentative satellites are depleted (\textsc{arg} share $-.23$, $p=.001$; \textsc{evaluation} $-.14$, $p=.011$), one relation dominates the tree (max-relation share $+.21$, $p=.005$), and the junctions are implicitly marked (asyndeton .69 vs.\ .63, $p=.013$). The mechanism proposed there applies here as well, since the parser assigns \textsc{elaboration} or \textsc{joint} to the unmarked junctions where the operative relations are semantic ones such as \textsc{evidence} or \textsc{cause}, which very often hold through an unstated warrant, and the models observe the same unmarked surface. The committed minority of 71 consensus enthymemes are those whose argumentative structure is visible. Two trees of the uncommitted majority are shown in Figures~\ref{fig:rst_3283} and~\ref{fig:rst_1142} of Appendix~\ref{sec:appendix_rst_trees}.

\subsection{Summary of the Success and Error Analysis}
\label{sec:analysis_summary}

We conclude with three observations. First, the asymmetry of the poles has a direct reading, in that certain absence of an enthymeme is mostly learned while certain presence mostly is not. This asymmetry, like the commitment of betabin at the poles and its restraint where the readers divide (Section~\ref{sec:contested_profiles}), is measurable only through the variation metrics, and at no cost on F1; a model can be made to perform as well on the verdict while behaving better on the votes. The attribution--evaluation configuration of Section~\ref{sec:contested_profiles} shows moreover that the variation so captured is discourse-principled reader variation, and that a model can be taught to capture it. The per-vote consistency metric completes this picture on the other side, since it sheds light on the inconsistent votes and represents them with more precision than any of the other variation metrics (Section~\ref{sec:consistency_profiles}).

Second, the two failure profiles reverse the two easy profiles almost feature by feature, with the sign flipping between the columns on spine ratio, argumentative share, \textsc{evaluation}, \textsc{explanation}, the judgment configurations and max-relation share. Success and failure are thus outcomes of the same reading, in which the models rely on surface argumentative structure as a proxy for the label; the proxy is reliable exactly when discourse form and label align, and when the two dissociate the error runs in both directions at once, moving the none that argues explicitly toward the enthymeme region and leaving the enthymeme without argumentative marking near the threshold. This last point can be explained but the fact the implicit content itself, the object the annotators were asked to identify, is not what any model reads, and at the unmarked junctions where it operates, the parser makes the corresponding mistake, reading continuation where the relation is evidential (Section~\ref{sec:evidence_elaboration}).

Third, the failure appears structural, since the enthymeme-side set spans 81\% of its band including nearly the whole unanimous band, and the learning curves of Section~\ref{sec:learning_curves} give no indication that any metric family improves onto these tweets with more data.

\section{Discourse Structure Inside the Model}
\label{sec:probing}

The injection results of Section~\ref{sec:results} showed that RST features are redundant for the fine-tuned models, and the error analysis of Section~\ref{sec:error_cases} showed that their successes and failures follow discourse profiles the features describe. This section asks whether these two observations have a common representational basis, namely whether the discourse dimensions of the profiles are encoded inside the model, and whether fine-tuning for enthymeme detection promotes or sheds them.

\subsection{Probing Setup}
\label{sec:probing_setup}

We compare the pretrained DeBERTa encoder with the same encoder after enthymeme fine-tuning, both frozen, and represent each tweet by its last-layer CLS state.\footnote{Layers 0, 3, 6, 9 and 12 were probed; fine-tuning leaves the lower layers essentially unchanged, so we report the last layer, where the task reorganizes the encoder. Fine-tuned weights are those of the text-only soft model.} The probed concepts are binary discourse configurations derived from the profile dimensions,\footnote{We do not probe the 24 dimensions of Section~\ref{sec:features} as they stand, since the probing contrast requires binary concepts, whereas most dimensions are ratios; continuous dimensions were binarized or combined into the configurations listed here. The selection follows the discourse profiles of Section~\ref{sec:error_cases}, keeping the battery to the configurations relevant to the argument of this paper.} covering single relations (\textsc{evaluation}, \textsc{attribution}, \textsc{elaboration}), configurations (\textsc{contrast}$\,\wedge\,$\textsc{evaluation}, \textsc{attribution}$\,\wedge\,$\textsc{explanation}, embedded \textsc{joint}, high asyndeton), a disjunction of stance configurations, contestedness on compact self-contained trees, and the gold trope flags. Each concept is probed by a logistic regression on the frozen representation, scored by out-of-fold macro-F1, and the quantity of interest is the fine-tuning shift $\Delta$, the probe score on the fine-tuned encoder minus the score on the pretrained one. A positive $\Delta$ means the task promotes the concept in the representation, a negative one that it sheds it.\footnote{Eight random binary signatures calibrate the probe noise, with shifts within $\pm.038$ (mean $.011$), so only shifts of about $+.04$ and above count as promotion; the label itself gains $+.084$, the ceiling any concept could approach.}

\subsection{Probing Results}
\label{sec:probing_results}

\begin{figure}[!tbp]
    \centering
    \includegraphics[width=0.62\textwidth]{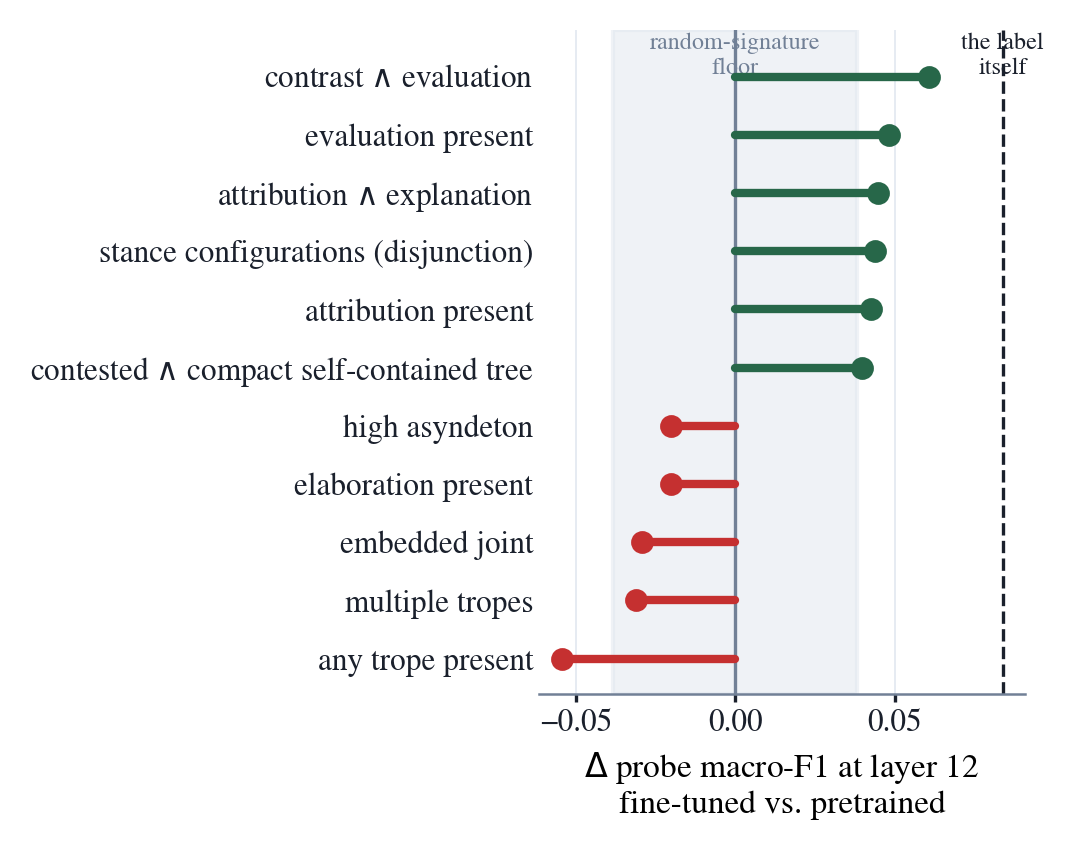}
    \caption{Fine-tuning shift $\Delta$ per probed signature (probe macro-F1 on the fine-tuned encoder minus the pretrained one, last-layer CLS). The shaded band marks the random-signature floor ($\pm.038$); the dashed line marks the label's own shift ($+.084$).}
    \label{fig:probe_signatures}
\end{figure}

As we can see from (Figure~\ref{fig:probe_signatures}), the main observation is that Fine-tuning promotes the scaffold of explicit argument. Every promoted configuration is a stance or attribution configuration, with \textsc{contrast}$\,\wedge\,$\textsc{evaluation} at $+.061$, \textsc{evaluation} at $+.048$, \textsc{attribution}$\,\wedge\,$\textsc{explanation} at $+.045$, the stance disjunction at $+.044$ and \textsc{attribution} at $+.043$. These are the configurations of the easy-enthymeme profile (Table~\ref{tab:easy_profile}) and the contestedness cue of the \textit{where} analysis, so the probing recovers inside the encoder the same proxy the error analysis recovered from the outputs. Contested tweets with compact, self-contained trees are also promoted ($+.040$), and selectively so, since disagreement accompanied by complexity cues shifts only $+.017$ and contestedness in general not at all ($-.001$). The encoder learns to recognize the tweets whose disagreement has no structural excuse.

Continuation structure and implicit marking, by contrast, are not promoted. \textsc{elaboration} ($-.020$), embedded \textsc{joint} ($-.029$) and high asyndeton ($-.020$) all shift slightly downward, but within the random-signature floor, so the task does not strengthen them. Fine-tuning thus leaves aside the surface under which the unstated warrants of the failure analysis sit, and the model does not become more attuned to the unmarked junctions where the annotators locate the enthymemes it misses.\footnote{This is consistent with the failure profiles, since the uncommitted consensus enthymemes are high-asyndeton, single-relation trees, and a representation that does not strengthen implicit marking cannot separate them from plain continuation.}

Tropes move in the opposite direction from discourse structure. While the discourse configurations are promoted, trope content is shed. The presence of any trope drops by $-.055$, the largest demotion of the battery and the only one beyond the random-signature floor, and multiple tropes ($-.031$) and distrust of experts ($-.019$) move in the same direction within it.\footnote{The trope flags are gold annotations, so the drop is not parser noise; the fine-tuned representation genuinely retains less content-level information than the pretrained one.} Enthymeme training thus recruits discourse form and discards content, which is consistent with the injection results of Section~\ref{sec:results}, where the RST features were redundant but stable while the trope oracle failed to transfer. Effect sizes are moderate and the battery is a screening rather than a hypothesis test,\footnote{Single training seed.} so we read the pattern rather than the individual numbers. We can conclude overall that fine-tuning promotes the explicit argumentative configurations, leaves implicit marking and continuation unpromoted, and sheds content.

\section{Discussion}
\label{sec:discussion}

This section steps back from the models to ask what our results mean in practice: what is lost when systems that classify and moderate social media content are trained and evaluated on majority labels, and whether evaluating against the variation among readers can help. We saw what evaluating under F1 alone can hide from us. On the basis of our results, we can only speculate on what such pipelines miss, but the profiles above indicate what a verdict-based evaluation leaves unmeasured. 

We wish nonetheless to point out that the aggregation practice follows a known platform logic: content ranking and moderation at scale rest on aggregated signals, quantified engagement on one side \cite{vandijck2014datafication,zuboff2019age}, and human judgments collapsed into majority labels for training, a practice shown to hide disagreement in social-computing tasks \cite{gordon2021disagreement} and observed in a deployed moderation classifier \cite{gordon2022jury}. Such pipelines are therefore blind to a linguistic reality in which disagreement and semantic interpretation are constitutive parts of the phenomena being classified, enthymemes among them, which is one of the central critiques advanced by perspectivist NLP \cite{aroyo2015truth,basile-etal-2021-need,plank-2022-problem,cabitza2023perspectivist}.

The stakes for the distortion of discourse ecosystem are significant for the controversial implicit content that goes undetected. Indeed, implicit content is accommodated into the common ground \cite{lewis1979scorekeeping,stalnaker2002common}, and what is accommodated repeatedly over time sediments into the taken-for-granted knowledge of a community \cite{berger1966social,lombardivallauri2016exaptation}. Hence a missed enthymeme could potentially be a premise left to sediment into commonsense, which is how controversial content polarizes without ever arguing: communities come to differ not in the claims they debate but in the premises they no longer state.

We take up this point by returning to what the variation metrics make visible, and by speculating on how they could help mitigate such distortions for implicit controversial content.

\subsection{Closing the Case for F1 in Semantic Interpretation Tasks}
\label{sec:f1_case}

But first we ought to give back to F1 what belongs to F1. The missed paratactic enthymemes of Section~\ref{sec:error_cases}, whose clauses sit side by side with no marked link, are partly visible to the verdict metric, since those that fall below the threshold count as errors, and a better model under F1 would mitigate part of the risk of distortion mentioned above; on our data it already catches the enthymemes whose argumentative satellites are visible and clears the flat single-relation nones. 

However, our data gives little reason to expect much more. Every transformer configuration we trained lands between 0.65 and 0.69 macro-F1, the learning curves are flat from 300 examples on, and the failure on the paratactic profile spans 81\% of the consensus enthymemes with no sign of improving with data. This is consistent with what is reported on other semantic interpretation tasks where annotators genuinely diverge: on natural language inference, models reach near-perfect accuracy on high-agreement items while barely beating chance on the contested ones, which questions the value of further optimizing hard-label metrics on that portion of the data \cite{pavlick-kwiatkowski-2019-inherent,nie-etal-2020-learn}.

Furthermore, a deployed detector creates an adversarial scenario in which producers adapt the surface until the metric is flunked, as with AI-text detectors defeated by paraphrasing and humanization \cite{krishna2023paraphrasing,li2026beyond}, and in our case the paratactic profile, such as the one in Figure~\ref{fig:rst_1142}, shows exactly which surface to adopt. If the verdict cannot be predicted much better because it collapses a divided judgment, and if optimizing it only redirects content toward the form it misses, then F1 is not the adapted metric for this content, and the question becomes what should be measured instead.

\subsection{Variation Metrics and the Discourse Ecosystem}
\label{sec:ecosystem}

With the lens of our variation metrics we can speculate on several discourse ecosystem distortions. What these metrics observe, and what F1 does not, is the difference between a model that decides, a model that hedges, and a model that registers a split. F1 turns every probability into a verdict, so a prediction obtained at the threshold and a committed one enter ranking and moderation pipelines with the same authority. The variation metrics are distributional, meaning they keep these behaviors apart, and each of the distortions below arises from erasing one of these differences before the score reaches the systems that act on it.

A first distortion concerns the consensus nones on which the models remain at the
threshold, which the error analysis of Section~\ref{sec:pole_failures} identifies as tweets
that state a complete argument, a judgment supported by an explicit reason. F1 counts a probability of .47 on a 0/5 tweet
as a correct rejection. The depth and calibration metrics show instead that the model never
decided, as Figure~\ref{fig:none_pole_sets} illustrates. In a moderation or ranking pipeline
that reads the raw score, these tweets enter with a near-controversial score although every annotator rejected them. What raises the score is the discourse
form of these tweets, a judgment supported by an explicit reason, not any implicit claim
they make. The distortion is therefore a bias against explicit argumentation: content that
states its premises receives the same treatment as content that leaves them unstated, and a
pipeline that demotes or flags near-boundary scores acts on the former as if it were the
latter. A calibrated distributional output separates the confident rejection from the
undecided one, and the betabin results show that this separation is obtained at no cost on
F1.

A second distortion concerns the contested tweets of Section~\ref{sec:success_analysis}, whose reported speech from an exterior source together with a judgment of it divides the annotators. Here F1 works against the ecosystem directly, since it penalizes the model that returns these tweets to .51, where the annotators actually place them, as exemplified in Figure~\ref{fig:where_pole}, and rewards the model that settles a reading no reader consensus supports, attributing quoted content to the tweet author or resolving irony one way for all readers. A model evaluated on the \textit{where} metric can instead flag these tweets as contested, so that a pipeline can reroute them and withhold automatic action on them rather than enforce one reading.

A third distortion concerns the consensus enthymeme pole of Section~\ref{sec:error_cases}, on which no model commits and threshold tuning converts near-threshold scores into verdicts. This case the variation metrics cannot repair, since the certain presence of an enthymeme is not learned by any of our systems; what they can do is make it honest, in that a pipeline reading the distribution knows that these verdicts were never backed by commitment, which separates a declared blind spot from a false clearance.

\section{Conclusion}
\label{sec:conclusion}

This paper set out to detect enthymemes in tweets and to account for the variation among the readers who judge them. We close by returning to what each part of the study established, and on what evidence.

On detection, the result is a ceiling on macro-F1 that no configuration we trained exceeds. Every fine-tuned configuration, under hard and soft labels, with and without injected features, and with a vote-count head, scores between 0.65 and 0.69 macro-F1 (the hard-label model once its threshold is tuned), and the feature-only and TF-IDF classifiers stay below this range (Table~\ref{tab:test_results}). The learning curves of Section~\ref{sec:learning_curves} are flat from 300 examples on (Figure~\ref{fig:learning_curve}), so neither more data nor a different objective moves the verdict. We read this as a property of the task rather than of the models. Enthymeme detection is a semantic interpretation task on which five readers disagree on 37\% of items, and Figure~\ref{fig:f1_blindspot} shows the consequence for the best system, whose probabilities all fall between 0.07 and 0.62 and whose contested bands are indistinguishable from each other. Against this ceiling we introduced betabin (Section~\ref{sec:betabin}), which predicts a distribution over the vote count instead of a label. It scores at parity with the text-only baseline on macro-F1 and is the only non-baseline configuration that never loses to it.

On evaluation, the paper replaced the verdict with a set of questions about the vote distribution (Section~\ref{sec:existing_metrics}): where the contested tweets are, how far the probabilities sit from the vote fractions, whether they are calibrated in vote units, and, with the per-vote metric of Section~\ref{sec:pervote}, whether the model's errors fall on the votes that are unusual for the annotator who cast them. Under these metrics the systems that tie on F1 separate (Table~\ref{tab:soft_metrics}), and the properties they measure improve with data where the verdict does not (Figure~\ref{fig:soft_metrics_curve}). Betabin obtains its advantage through two behaviors. It moves the contested tweets of the \textit{where} set back to .51, where the annotators place them (Figure~\ref{fig:where_pole}), and it commits on the unanimous nones that the text-only model holds at .47 (Figure~\ref{fig:none_pole_sets}). Both behaviors are invisible to a metric computed against the majority label.

The per-vote metric adds a distinction the distributional metrics do not make. It scores each vote against the behavioral profile of its annotator, so that a dissenting vote which is routine for one annotator and unusual for another is not treated as the same event. The tweets on which this metric decides the ordering are unanimous nones framed as rhetorical questions, the \textsc{topic--comment} configuration of Section~\ref{sec:consistency_profiles}, on which one annotator can build a full argument and still vote \textit{none} (Figure~\ref{fig:rst_481}). Since a vote that departs from its annotator's otherwise stable behavior is the one most likely to be inconsistent, modeling the votes and conditioning on annotator behavior gives a first step toward separating inconsistent annotation from genuine disagreement, on the evidence of the vote matrix alone.

On discourse structure, the paper found the RST features redundant for the fine-tuned models (Section~\ref{sec:results}) and then explained why. The probing of Section~\ref{sec:probing} shows that fine-tuning promotes the explicit stance and attribution configurations but not \textsc{elaboration}, embedded \textsc{joint} or asyndeton (Figure~\ref{fig:probe_signatures}). The encoder already represents what the features encode. The same dimensions, used descriptively, show that the models rely on explicit argumentative structure as a proxy for the label. The proxy holds on the easy tweets of Table~\ref{tab:easy_profile}, where the easy none is a sequence of nuclei under one relation and the easy enthymeme carries argumentative satellites close to the root. It fails when form and label dissociate, and it fails in both directions at once (Table~\ref{tab:failure_two_profiles}). The nones that state a judgment and support it are held at the threshold, and the enthymemes without argumentative marking are never committed.

The second of these failures is the finding we consider most consequential. It concerns 81\% of the consensus enthymemes, and the learning curves give no indication that it recedes with data. These tweets share a profile of implicitly marked junctions, a single dominant relation and little reported speech (Section~\ref{sec:where_failures}, Table~\ref{tab:where_failure_profile}). At the unmarked junctions the parser assigns \textsc{elaboration} or \textsc{joint}, an additive relation in the sense of \citet{sanders1992toward}, where the operative relation is \textsc{evidence} or \textsc{cause}, a pragmatic one that holds through an unstated warrant. Figure~\ref{fig:evidence_elaboration} shows the two construals on one tweet, and Figures~\ref{fig:rst_3283} and~\ref{fig:rst_1142} show trees on which five annotators agreed and the parser recorded no argumentative relation at all. What distinguishes the two readings is not in the text. It is the warrant a reader supplies, and the parser, like the detection system, supplies none. The error analysis therefore identifies a failure of the parser and of the systems together, on the same surface, and it locates the limit of current detection at the certain presence of a component that the text does not contain.

The contested middle has a discourse profile of its own. The tweets on which readers divide combine an \textsc{attribution} node with an \textsc{evaluation} satellite, reported speech from an exterior source together with a judgment of it, and the judgment is often ironic or a rhetorical question (Section~\ref{sec:contested_profiles}, Figure~\ref{fig:rst_1965}). The configuration produces both kinds of disagreement identified by \citet{pastor-vanarkel-2026-disagreement}, costly resolution of the exterior source and subjective perception of the speech act. Betabin learned that this configuration yields split votes, and the probing shows that fine-tuning promotes the same configuration inside the encoder. Interpretive variation on this task has a discourse-structural basis, and a model trained on the votes can learn it.

The consequences of evaluating on the verdict alone extend beyond the models. A premise left unstated and accepted sediments into what a community no longer argues about (Section~\ref{sec:discussion}), and a detector evaluated on the verdict lets it pass on the consensus enthymemes and settles by a threshold the tweets on which readers divide (Section~\ref{sec:ecosystem}). A pipeline built on the alternative can withhold action where readers divide and can leave the argument that states its premises out of the near-controversial band. What such a pipeline needs is not more text but reader interpretation. The variation measured here exists only because five people judged each tweet, and the content the systems miss is supplied by the reader's background, not by the text (Section~\ref{sec:evidence_elaboration}). Systems trained on a majority label fix what counts as controversial at the moment of annotation and remove the variation that could revise it. Systems that keep reader judgments, keep collecting them and representing their variation, leave that question open and able to change. The metrics and the model introduced here are expected to contribute to the second kind, in which what a community treats as settled remains visible as a distribution of readings rather than a verdict. The premise that passes unmeasured may be one of the ways polarization takes hold, and a pipeline of this kind is at least able to see it.

These results point to two directions for future work. First, more data of the same kind is unlikely to help, since the learning curves are flat. What should grow instead is the information collected per item, with more annotators per tweet, individual votes rather than a majority label, and each annotator's reconstruction of the missing component, so that models can learn the inference readers make and not only its outcome. Second, the parser's blind spot at unmarked junctions calls for discourse parsers that can recover implicit relations such as evidence, and for manually corrected trees against which to test them.

\FloatBarrier
\section{Limitations}
\label{sec:limitations}

A limitation of the research presented here is the size of the dataset, which provides limited insights into the types of patterns we observe, particularly with regard to model generalisation—even within the same type and topic distribution. This leads us to note that the phenomena we identify may be closely tied to our specific topic.

Moreover, it is important to note that we only work with generated data for RST annotations, and that better labeling of less recognized relations by parsers could have yielded different results.

\section*{Acknowledgments}
This work was produced as part of the HYBRIDS project, a Marie Sk{\l}o\-dow\-ska-Curie Doctoral Network funded by the European Union under grant no.\ 101073351 and the UK Research and Innovation (UKRI) Horizon Funding Guarantee, and the AI-CODE project, funded under the European Union's Horizon Europe research and innovation programme, grant agreement no.\ 101135437. We also acknowledge the annotators: Nienke Gelderland, Afrooz Bolboli, Jarno Van Arkel, Peter Beinema, Ulas Askin, Huixin Lan and Pascale Mannone.

\FloatBarrier
\bibliographystyle{compling}

\bibliography{custom}

\appendix
\counterwithin{figure}{section}
\counterwithin{table}{section}

\section{RST Feature Encoding}
\label{sec:appendix}

\begin{table}[H]
\centering
\caption{The 24-dimensional RST feature encoding. All features are indicators (present or absent) or ratios computed over the automatically generated RST tree. The candidate features were constructed from two motivated sources: the argumentative-satellite patterns hypothesized to distinguish explicit from implicit arguments, and the discourse configurations associated with annotator disagreement in our data. From this pool, we kept the 24 dimensions that remained predictive across training folds. We kept features given nonzero weight by an L1 logistic regression in at least three of five training folds.}
\label{tab:v9_features}
\scriptsize
\begin{tabularx}{\textwidth}{@{} c >{\raggedright\arraybackslash}X >{\raggedright\arraybackslash}X @{}}
\toprule
\textbf{Dim} & \textbf{Feature} & \textbf{Motivation} \\
\midrule
\multicolumn{3}{c}{\textit{Relational composition}} \\
\midrule
1--6 & Satellite of each core relation present at depth $\leq 2$ (indicators) & Discourse relations are the feature type that helped most in \citeauthor{chernyavskiy-etal-2024-unleashing} (\citeyear{chernyavskiy-etal-2024-unleashing}); a satellite with one of these relations gives explicit support to the main claim. \\
7 & Share of core-relation satellites among all satellites & How much of the extra information supports the claim rather than just adding to it. \\
8 & Character span of the shallowest core-relation satellite / text length & How much of the tweet is spent on explicit support. \\
9 & Fraction of core-relation satellites in SN order & Whether the support comes before the claim. \\
10 & Spine ratio: EDUs on an all-nucleus path / number of EDUs & How much of the tweet is essential content (Nucleus) with no extra information attached. \\
11 & NN nodes / all internal nodes & Whether the tweet joins statements of equal weight or subordinates one to another; nuclearity was the other most useful feature type in \citeauthor{chernyavskiy-etal-2024-unleashing} (\citeyear{chernyavskiy-etal-2024-unleashing}). \\
\midrule
\multicolumn{3}{c}{\textit{Distributional and configurational}} \\
\midrule
12 & \textsc{elaboration} share of internal nodes & How much of the tree just adds detail. \\
13 & Dominant-relation share (most frequent relation / internal nodes) & Whether one single relation organizes the whole tweet. \\
14 & Relation diversity / internal nodes & How varied the relations are, independent of tree size. \\
15 & \textsc{topic-comment} present (indicator) & A framing pattern frequent in social media posts. \\
16 & Asyndeton: share of adjacent EDUs joined with no connective & Statements placed side by side with nothing marking the link; the reader must supply it. \\
17--18 & \textsc{joint} at the root; \textsc{joint} under a non-joint root (indicators) & Whether the tweet is a list of claims, or accumulates claims inside a larger structure. \\
19--20 & \textsc{attribution} share of internal nodes; \textsc{attribution} present & Reported speech: the tweet says who said what. \\
21 & Judgment relation (\textsc{evaluation} or \textsc{contrast}) present with no reason relation & A judgment with no stated grounds. \\
22 & Judgment and reason relation both present & A judgment with its grounds stated. \\
23--24 & NS share; SN share of internal nodes & In which direction extra information attaches across the tree. \\
\bottomrule
\end{tabularx}
\end{table}

\FloatBarrier
\section{RST Trees of the Discourse Profiles}
\label{sec:appendix_rst_trees}

\begin{figure}[!tbp]
\centering
\includegraphics[width=\textwidth]{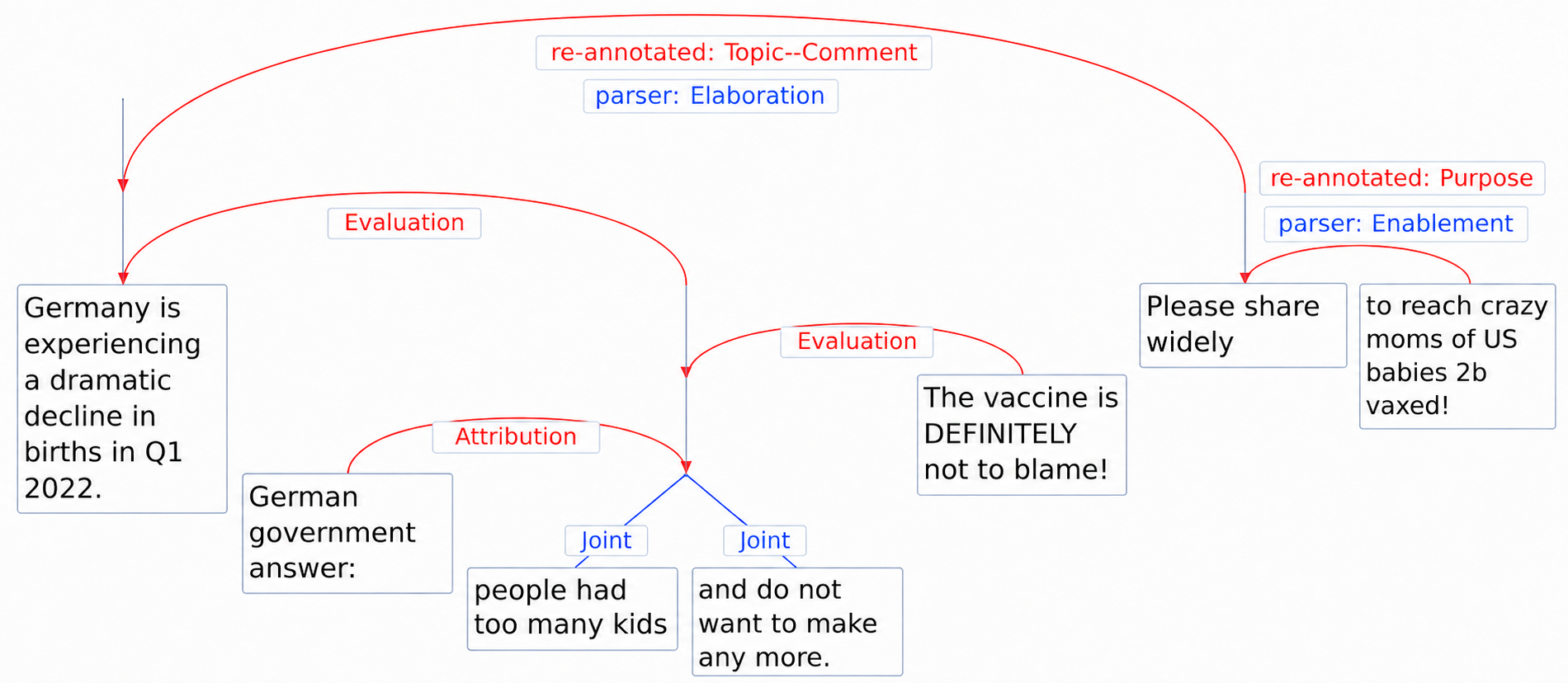}\\[6pt]
\begin{minipage}{0.96\textwidth}
  \small
  \noindent\textbf{Tweet \#1965} (2/5 enthymeme votes; \textit{where} success set)\\[2pt]
  \textit{``Germany is experiencing a dramatic decline in births in Q1 2022. German government answer: people had too many kids and do not want to make any more. The vaccine is DEFINITELY not to blame! Please share widely to reach crazy moms of US babies 2b vaxed!''}\\[6pt]
  \begin{tabularx}{\linewidth}{@{}l X@{}}
    $P_1$: & Germany is experiencing a dramatic decline in births in Q1 2022 \\[3pt]
  $P_2$: & \textcolor{blue}{Dramatic birth declines in Germany are vaccine-related \textit{(implicit)}} \\[3pt]
  \multicolumn{2}{@{}l@{}}{\hrulefill} \\[3pt]
  $C$: & The vaccine is DEFINITELY not to blame! [ironic] \\[3pt]
  \end{tabularx}
\end{minipage}
\caption{Contested \textit{where}-success set (Section~\ref{sec:contested_profiles}), 2/5 enthymeme votes, reconstruction of annotator 1. The backbone of the profile is visible: an \textsc{attribution} node (``German government answer:'') reports an explanation from an exterior source, and an \textsc{evaluation} satellite (``The vaccine is DEFINITELY not to blame!'') judges it ironically. The conclusion is explicit only through the irony, and the annotators split three ways (premise, conclusion, none). Where a manual re-annotation departs from the parser, both labels are shown (re-annotated above, parser below).}
\label{fig:rst_1965}
\end{figure}

\begin{figure}[!tbp]
\centering
\includegraphics[width=\textwidth]{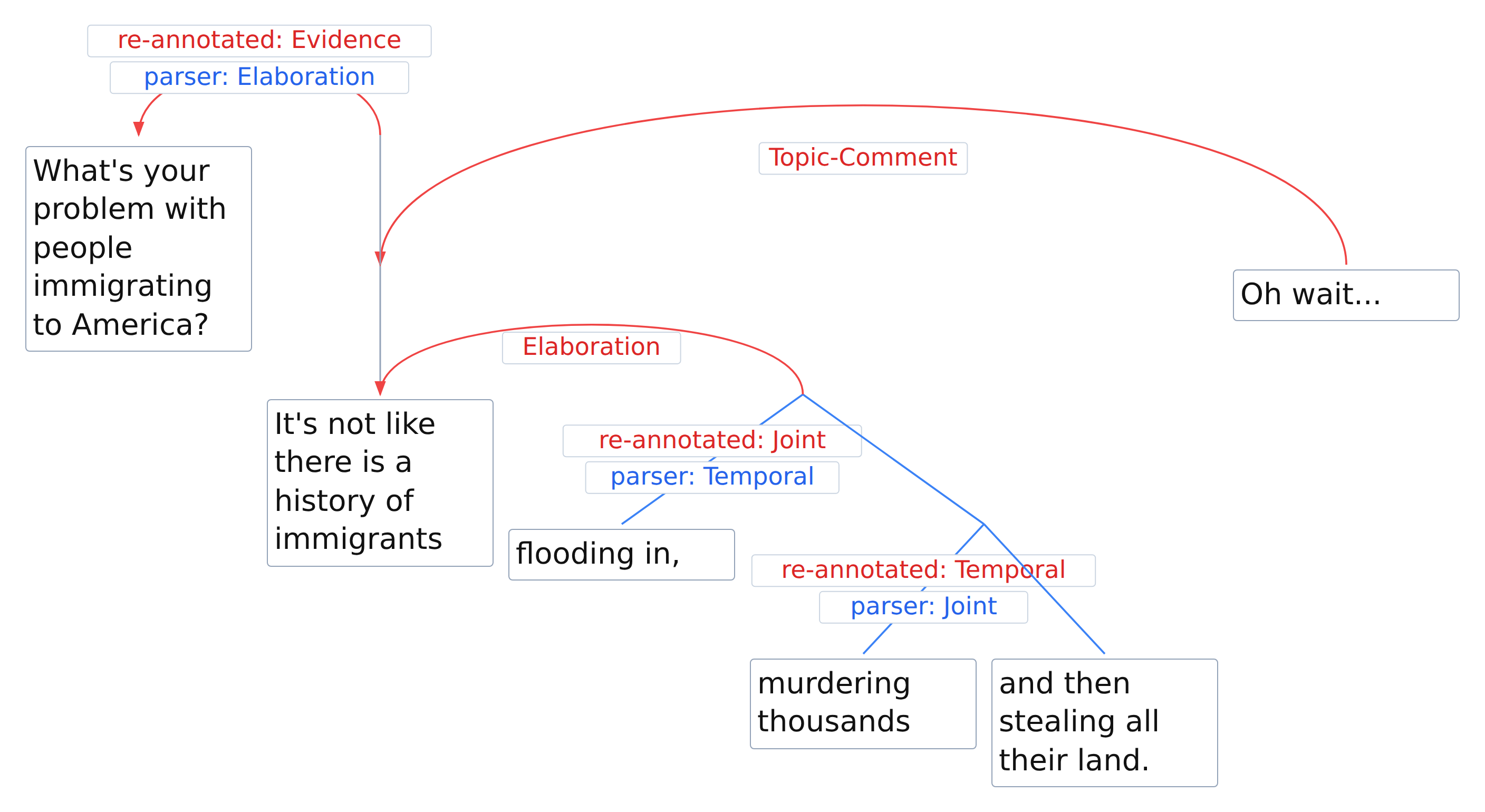}\\[6pt]
\begin{minipage}{0.96\textwidth}
  \small
  \noindent\textbf{Tweet \#3283} (5/5 enthymeme votes; uncommitted consensus enthymemes)\\[2pt]
  \textit{``What's your problem with people immigrating to America? It's not like there is a history of immigrants flooding in, murdering thousands and then stealing all their land. Oh wait...''}\\[6pt]
  \begin{tabularx}{\linewidth}{@{}l X@{}}
    $P_1$: & It's not like there is a history of immigrants flooding in, murdering thousands and then stealing all their land. Oh wait... [sarcastic: there is such a history] \\[3pt]
  $P_2$: & \textcolor{blue}{People whose ancestors engaged in mass immigration that displaced the original inhabitants have no moral standing to oppose current immigration \textit{(implicit)}} \\[3pt]
  \multicolumn{2}{@{}l@{}}{\hrulefill} \\[3pt]
  $C$: & What's your problem with people immigrating to America? [rhetorical: you have no grounds for one] \\[3pt]
  \end{tabularx}
\end{minipage}
\caption{Uncommitted consensus enthymeme (Section~\ref{sec:pole_failures}), 5/5 votes, implicit premise of annotator 1. Parser tree with manual corrections: where the re-annotation departs from the parser, both labels are shown (re-annotated in red above, parser in blue below), and the question's two EDUs are merged. The parser tree fits the profile of the set: no argumentative satellite, \textsc{elaboration} on half the nodes, unmarked junctions. It labels the second sentence \textsc{elaboration} where it is the evidence for the rhetorical question, and attaches the sarcasm marker ``Oh wait...'' to the whole tweet; the warrant linking evidence and question is the implicit premise, and nothing in the parser's tree encodes it.}
\label{fig:rst_3283}
\end{figure}

\begin{figure}[!tbp]
\centering
\begin{minipage}[c]{0.60\textwidth}
  \centering
  \includegraphics[width=\linewidth]{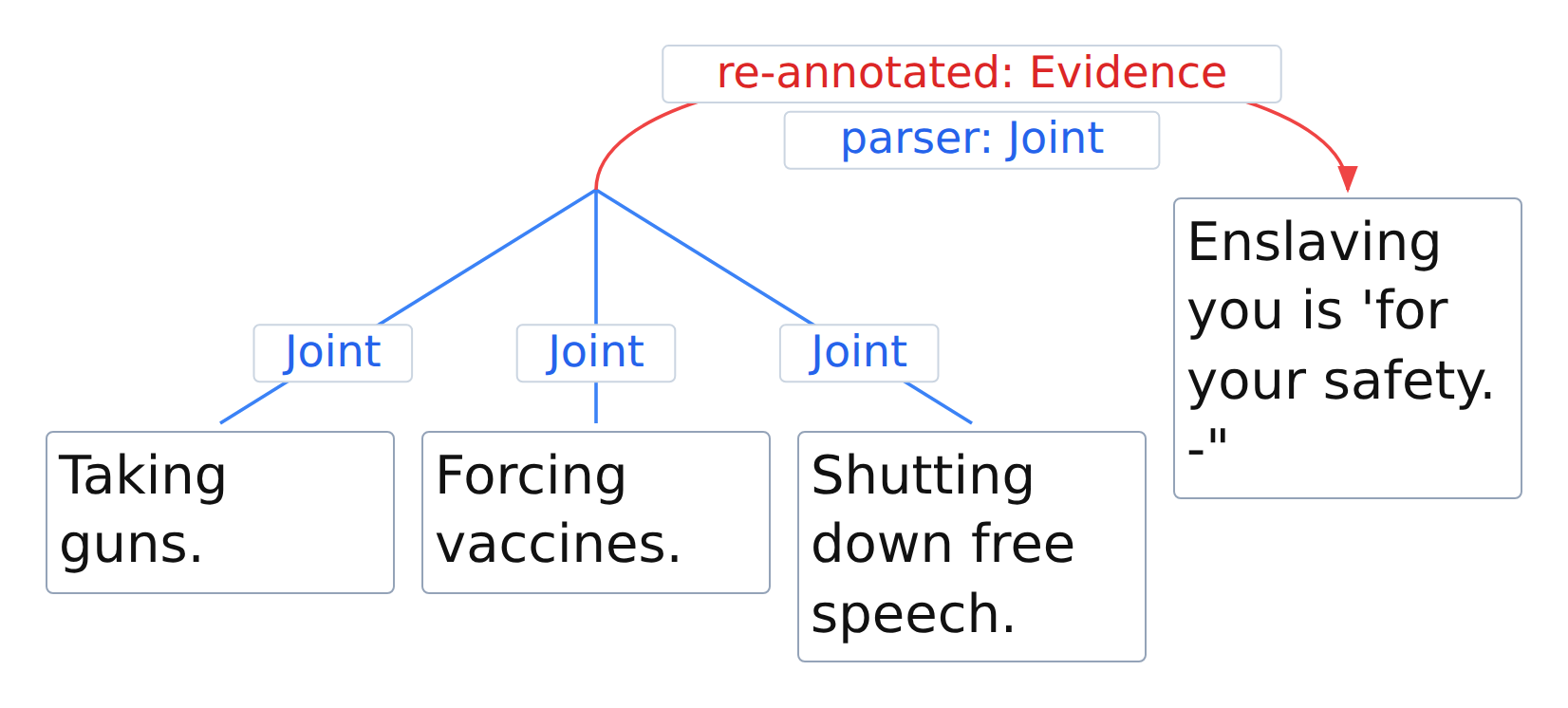}
\end{minipage}\hfill
\begin{minipage}[c]{0.38\textwidth}
  \small
  \noindent\textbf{Tweet \#1142} (5/5 enthymeme votes; uncommitted consensus enthymemes)\\[2pt]
  \textit{``Taking guns. Forcing vaccines. Shutting down free speech. Enslaving you is 'for your safety. -''''}\\[6pt]
  \begin{tabularx}{\linewidth}{@{}l X@{}}
    $P_1$: & \textcolor{blue}{Taking guns, forcing vaccines and shutting down free speech is a form of enslavement \textit{(implicit)}} \\[3pt]
  $P_2$: & [they are] Taking guns. Forcing vaccines. Shutting down free speech \\[3pt]
  \multicolumn{2}{@{}l@{}}{\hrulefill} \\[3pt]
  $C$: & Enslaving you is 'for your safety. -'' \\[3pt]
  \end{tabularx}
\end{minipage}
\caption{Uncommitted consensus enthymeme (Section~\ref{sec:pole_failures}), 5/5 enthymeme votes. Parser tree with manual corrections (re-annotated in red above, parser in blue below; the list is flattened and the split last clause merged). The parser gives a bare paratactic \textsc{joint} chain with a spine ratio of 1 and no connective: the list of actions is the evidence for the judgment in the last EDU, but nothing in the parser's tree marks it as such.}
\label{fig:rst_1142}
\end{figure}

\begin{figure}[!tbp]
\centering
\includegraphics[width=\textwidth]{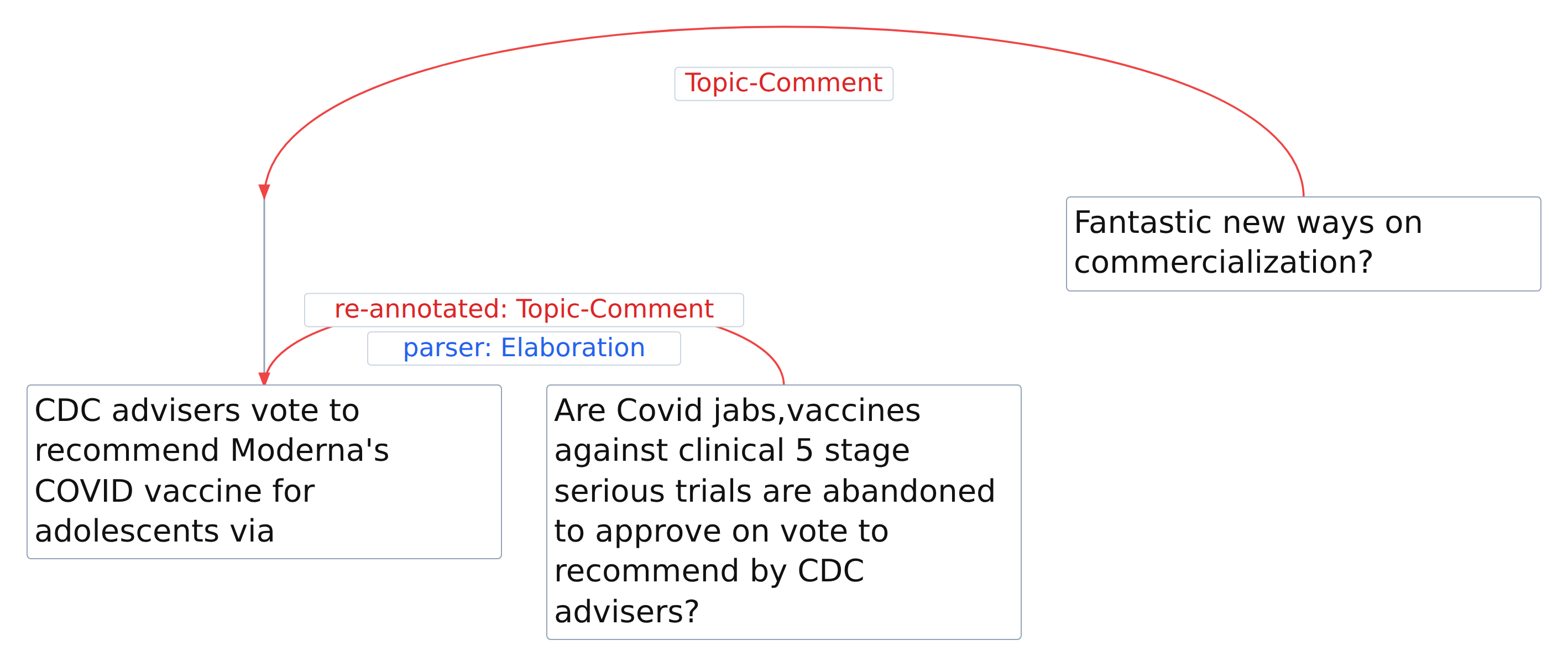}\\[6pt]
\begin{minipage}{0.96\textwidth}
  \small
  \noindent\textbf{Tweet \#481} (0/5 enthymeme votes; consistency set)\\[2pt]
  \textit{``CDC advisers vote to recommend Moderna's COVID vaccine for adolescents via Are Covid jabs,vaccines against clinical 5 stage serious trials are abandoned to approve on vote to recommend by CDC advisers? Fantastic new ways on commercialization?''}
\end{minipage}
\caption{Consistency set (Section~\ref{sec:consistency_profiles}), 0/5 enthymeme votes. Parser tree with corrections (re-annotated in red above, parser in blue below; the split first question is merged). Both \textsc{topic--comment} satellites are rhetorical questions, which can be rephrased as propositions and used in an argument, a reconstruction step that tends to produce inconsistent annotator behavior: here one annotator built the full argument with a scheme and still voted \textit{none}. The text-only model only senses that this structure may yield ambiguous votes and stays near the threshold; betabin has learned that an enthymeme reading is possible but most likely an inconsistency, and commits the tweet toward \textit{none}. Only the per-vote consistency metric can observe this difference.}
\label{fig:rst_481}
\end{figure}

This appendix shows, for the discourse profiles of Section~\ref{sec:analysis} on which the models divide, one or two tweets whose tree exemplifies the profile. Each figure gives the RST tree of the full tweet on which the 24 dimensions of Table~\ref{tab:v9_features} were computed, rendered with the RST Annotator tool: satellites point to their nucleus with a red arc labeled by the relation, and multinuclear relations are in blue. The trees are the parser's output with manual corrections in the coarse RST-DT relation set: where a corrected relation departs from the parser's, both labels are shown, the re-annotated one in red above and the parser's in blue below, and mis-segmented EDUs are merged. Where the tweet is an enthymeme, the gold argument reconstruction follows in the format $P_1$, $P_2$, $C$, taken verbatim from the annotator whose decomposition is the soundest among those recorded for the tweet, with the implicit component in blue. Vote counts give the number of annotators, out of five, who labeled the tweet as an enthymeme. Figure~\ref{fig:rst_1965} shows the contested \textit{where}-success set of Section~\ref{sec:contested_profiles}; Figures~\ref{fig:rst_3283} and~\ref{fig:rst_1142} show the uncommitted consensus enthymemes of Section~\ref{sec:pole_failures}; Figure~\ref{fig:rst_481} shows the consistency set of Section~\ref{sec:consistency_profiles}.

\end{document}